\pdfoutput=1
\documentclass[11pt]{article}

\usepackage[preprint]{acl}

\usepackage{times}
\usepackage{latexsym}
\usepackage{algorithm, algpseudocode}
\usepackage{amsmath} 
\usepackage{tikz}
\usepackage{makecell}
\usetikzlibrary{arrows.meta,positioning,calc}
\usepackage{xurl}      % 支持长 URL 自动换行
\usepackage{hyperref}  % 自动加载 url，不再需要自己加载

\usepackage{booktabs, multirow, array, xcolor, colortbl, pifont}

\definecolor{mygreen}{rgb}{0.27,0.70,0.23}   % DEF1D3
\definecolor{myyellow}{rgb}{0.74,0.71,0.02}  % FDF4D1

\definecolor{cadmiumgreen}{rgb}{0.0, 0.42, 0.24}

\definecolor{myred}{rgb}{0.7, 0.3, 0.0}
\definecolor{myblue}{rgb}{0.2, 0.3, 0.6}

\usepackage[T1]{fontenc}
\usepackage[utf8]{inputenc}

\usepackage{microtype}

\usepackage{inconsolata}

\usepackage{graphicx}

\usepackage{booktabs, multirow} % for borders and merged ranges

\usepackage{tcolorbox} % for prompt design

\usepackage{hyperref}

\usepackage{longtable}

\usepackage[dvipsnames]{xcolor}

\title{NG-Router: Graph-Supervised Multi-Agent Collaboration \\ for Nutrition Question Answering}

\author{
\small
    Kaiwen Shi\textsuperscript{1}\textsuperscript{*},
    Zheyuan Zhang\textsuperscript{1}\textsuperscript{*}, 
    Zhengqing Yuan\textsuperscript{1},
    Keerthiram Murugesan\textsuperscript{3},\\
\small
    \textbf{
    Vincent Galassi\textsuperscript{1},
    Chuxu Zhang\textsuperscript{2},
    Yanfang Ye\textsuperscript{1}\textsuperscript{$\dagger$}} \\
\small
    \textsuperscript{1}University of Notre Dame, 
    \textsuperscript{2}University of Connecticut,
    \textsuperscript{3}IBM Research, \\
\small
    \textsuperscript{*}Equal Contribution 
    \textsuperscript{$\dagger$}Corresponding Author
    \\
\small
    \texttt{\{kshi3, yye7\}@nd.edu},
}

\begin{document}
\maketitle

\begin{abstract}
Diet plays a central role in human health. Nutrition Question Answering (QA) has emerged as a promising paradigm to deliver personalized dietary guidance and prevent diet-induced chronic conditions, yet despite recent progress, existing approaches struggle with two key limitations: 1) the limited capacity of single agents for domain-specific reasoning and the inherent design complexity of multi-agent systems; 2) the overload of contextual information that dilutes downstream decision-making. In this paper, we introduce \textbf{Nutritional-Graph Router (NG-Router)}, a framework that formulates nutritional QA as a supervised, knowledge-graph–guided multi-agent collaboration problem. Our approach integrates agent nodes into knowledge graphs and employs a graph neural network to learn task-aware routing distributions over agents, leveraging soft supervision derived from empirical agent performance. To further mitigate contextual overload, we incorporate a gradient-based subgraph retrieval mechanism that identifies salient evidence during training, thereby enhancing reasoning in multi-hop and relational settings. Extensive experiments across multiple benchmarks and backbones demonstrate that NG-Router consistently surpasses both single-agent and ensemble baselines, achieving robust and generalizable improvements. By embedding collaboration schemes directly into graph-supervised signals, our method offers a principled path toward domain-aware multi-agent reasoning for complex nutritional health tasks. Our code repo can be found \href{https://anonymous.4open.science/r/NGRouter-E7FB/}{here}.
\end{abstract}

\begin{figure*}[!t]
  \centering
  \includegraphics[width=\linewidth]{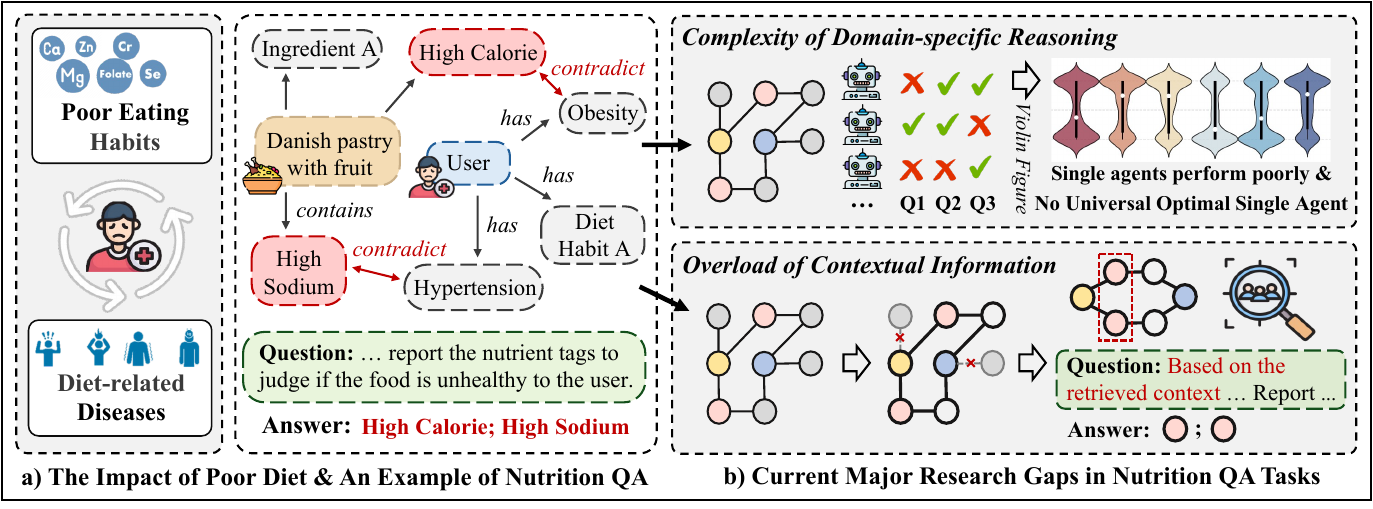}
  \vspace{-20pt}
  \caption{Illustration of Nutrition QA challenges. (a) Poor dietary habits lead to health risks and complex user–food–condition interactions, as shown in a personalized QA example. (b) Two key research gaps emerge: (i) domain-specific reasoning requires multiple complementary agents, with no single model performing optimally across queries; and (ii) excessive and unstructured contextual information hampers accurate retrieval and reasoning.}
  \vspace{-15pt}
  \label{fig:motivation}
\end{figure*}

\section{Introduction}

Diet is one of the most influential determinants of human health, shaping both disease prevention and long-term well-being. Yet, unhealthy eating remains widespread despite extensive public awareness of the benefits of proper nutrition \cite{WHO_2021_HealthyDiet}. In the United States, for instance, an estimated 42.4\% of adults are classified as obese \cite{CDC_2020_Obesity}. Globally, poor dietary patterns were associated with more than 11 million deaths in 2017, alongside millions of disability-adjusted life-years (DALYs), often linked to factors such as excessive sodium intake \cite{Afshin_2019_DietaryRisks, WHO_2023_Obesity}. These figures highlight the pressing need for large-scale interventions that foster healthier eating behaviors (Figure~\ref{fig:motivation} (a)). To address the challenges in personalized nutritional health, Nutrition Question Answering (QA) has gained traction as a practical solution due to its accessibility, low entry barrier, and interactive nature \cite{min2022applications, bondevik2024systematic}. Advances in LLMs further strengthen this direction by enabling more sophisticated reasoning over personalized dietary guidance \cite{zhang2025llms4all}. Although recent datasets and tasks \cite{bolz2023hummus, zhang2024ngqa} have advanced the field from a data-centric perspective, the effective use of these benchmarks remains constrained by two major research gaps:

\noindent\textbf{Complexity of Domain-Specific Reasoning.} Simple single-agent reasoning is usually insufficient for handling the medical and nutritional complexity required in personalized dietary guidance. While standalone LLM agents can perform well in general domains, growing evidence shows that coordinated systems of multiple agents often yield superior outcomes \citep{hongmetagpt, zhong2024debug, ma2025autodata}. Multi-agent frameworks (MAS) leverage diverse and complementary roles to improve reasoning accuracy, exploratory depth, and robustness \citep{shinn2023reflexion, qian2025scaling, wang2025mixture}. However, designing such systems is far from trivial. A pervasive challenge lies in determining which agents should participate in collaboration, as literature consistently shows that the performance of different agents and backbones varies substantially across tasks, and there is no universally optimal solution \cite{feng2024graphrouter}. This limitation becomes especially problematic in domains like healthcare, where domain knowledge is required for delicate reasoning (Figure~\ref{fig:motivation} (b)).

\noindent\textbf{Overload of Contextual Information.} Health-aware dietary reasoning typically involves an overwhelming volume of domain-specific information, such as medical conditions, nutritional profiles, food attributes, and condition-specific constraints \cite{zhang2024ngqa}. When this context is presented in full to an LLM or multi-agent system, the reasoning process can become diluted or misdirected, leading to inefficiency and factual errors \citep{jin2024graph, jiang2024kg, peng2024graph}. Effective personalized guidance therefore depends not only on reasoning ability but also on accurately retrieving and prioritizing the most relevant information from the broader context \cite{guo2024knowledgenavigator, wen2023mindmap}. Without mechanisms to filter, structure, and surface key evidence, model performance degrades, and the quality of generated answers suffers accordingly (Figure~\ref{fig:motivation} (b)).

To tackle the two aforementioned challenges, we adopt the Graph QA setting from the NGQA benchmark \cite{zhang2024ngqa} and introduce \textbf{Nutritional-Graph Router (NG-Router)}, a framework that casts nutritional question answering as a supervised, knowledge-graph guided multi-agent collaboration problem. To address the first challenge, we integrate agent nodes into the KG and design a heterogeneous graph neural network (GNN) that propagates information across node types and produces task-aware routing distributions over agents. Rather than assigning agents deterministically, the router leverages soft supervision derived from empirical agent performance to learn dynamic probability distributions, and it generates final answers via weighted aggregation of agent outputs. To address the second challenge, we extend the pipeline with a gradient-based subgraph retrieval mechanism. During training, gradient signals supervise the retrieval of salient subgraphs, effectively filtering away irrelevant context. This design is motivated by prior findings showing that such retrieval strategies are especially advantageous for complex multi-hop reasoning \citep{jin2024graph, jiang2024kg, peng2024graph}.

Overall, our method departs from heuristic voting or LLM-based judging by learning collaboration schemes directly from supervised graph signals. This enables the incorporation of domain structure into agent coordination without requiring handcrafted prior knowledge, while the gradient-guided retrieval substantially reduces contextual overload for downstream reasoning. Extensive experiments demonstrate that NG-Router consistently surpasses single-agent and ensemble baselines and generalizes well across both benchmarks and model backbones. Our contributions can be summarized as follows:

\begin{itemize}
\vspace{-5pt}
\item \textbf{KG–Driven Agent Collaboration.}
In this paper, we present a novel framework that converts multi-agent question answering into domain-specific knowledge graphs, where nodes represent not only queries and agents but also fine-grained entities and contextual interactions. We further train the graphs to learn adaptive collaboration strategies tailored to heterogeneous agent capabilities.
\vspace{-5pt}
\item \textbf{Graph-Supervised Subgraph Retrieval.}
To enhance downstream reasoning, we propose a retrieval mechanism that leverages gradient-based supervision during KG training. By identifying important nodes, the method extracts the most relevant subgraphs, enabling more precise and context-aware QA.
\vspace{-5pt}
\item \textbf{Extensive Empirical Validation.}
Comprehensive experiments in the Nutritional QA domain demonstrate that our collaboration framework consistently surpasses both the strongest single-agent models and state-of-the-art baselines across multiple tasks.

\end{itemize}

\section{Related Works}
\subsection{Task-Adaptive Agent Selection.}  
With the rapid development of LLMs and agentic frameworks, a growing body of work shows that no single model consistently dominates across tasks; instead, agents exhibit \emph{complementary strengths}. Multi-agent systems such as AgentVerse demonstrate collaborative gains over individual agents \citep{chen2024agentverse}, while ReConcile shows that organizing diverse LLMs into rounds of discussion with consensus voting improves reasoning \citep{chen2024reconcile}. These findings motivate \emph{input-conditioned selection and coordination}—i.e., routing or assembling specialized agents per query. Early efforts used static ensembling or binary collaboration (e.g., self-consistency \citep{wang2022self}) to exploit diversity, but these approaches predefine model sets and lack input adaptivity, offering limited guidance on agent prioritization \citep{jiang2023llm}. More recent work trains learned routers that select LLMs based on query characteristics. RouteLLM learns routing from human preference data \citep{ong2024routellm}, RouterDC uses dual contrastive learning to assemble multiple LLMs \citep{chen2024routerdc}, and MixLLM treats routing as a contextual bandit problem for dynamic adaptation \citep{wang2025mixllm}. Other methods focus on coordinating multiple agents rather than selecting a single one: TO-Router and BEST-Route determine which experts to involve based on query difficulty instead of fixed pipelines \citep{stripelis2024multi, ding2025best}. However, many of these approaches still rely on heuristic rules or shallow controllers and rarely model richer inter-dependencies among tasks, queries, and agents. A newer line of work frames routing as a structured learning problem \cite{zhang2025agentrouter}. \textsc{GraphRouter}, for example, formulates routing as link prediction on a heterogeneous graph and uses GNNs to model query--model and inter-model relations \citep{feng2024graphrouter}. Although this marks progress beyond simple heuristic methods , it still struggles to integrate fine-grained task semantics or supervised graph signals that more directly guide adaptive collaboration across diverse agent designs. More can be seen in Appendix~\ref{app:literature}

\subsection{Prior Works in Nutrition Personalization.}
With growing awareness of the importance of dietary health, various studies have sought to incorporate health metrics into applications such as food recommendation systems \cite{tian2022reciperec, tian2022recipe2vec, tian2021recipe}. These approaches can be grouped into three primary categories. First, some research emphasizes single indicators like calorie or fat content, as highlighted in works by Ge et al. \cite{ge2015health} and Shirai et al. \cite{shirai2021identifying}, though such metrics often fail to represent the multifaceted nature of a balanced diet. Second, simulated health data has been utilized, as demonstrated by Wang et al. \cite{wang2021market2dish}, but these methods often diverge from real-world data distributions. Finally, recent studies have applied global health guidelines to develop composite health scores, such as \cite{bolz2023hummus, zhang2024greenrec}. However, foods deemed healthy by general standards can still negatively affect certain individuals \cite{yue2021overview, zhang2024mopi}, highlighting the absence of a universal solution. Beyond this, there is yet an effective solution to address the overload contextual information this domain challenge brings.

\section{Preliminary}

\subsection{NGQA Benchmark Extension}
Nutritional Graph Question Answering (NGQA) benchmark is a set of graph-based question answering datasets for personalized nutritional health, constructed from the National Health and Nutrition Examination Survey (NHANES) and the Food and Nutrient Database for Dietary Studies (FNDDS) data, which evaluates whether a food is healthy for a specific user by linking medical conditions, dietary behaviors, and nutritional profiles. It supports a variety of reasoning tasks across varying question complexities and establishes a comprehensive evaluating system for nutritional question answering task. In NGQA benchmark, each question has a context knowledge graph $\mathcal{G}=(\mathcal{V},\mathcal{E})$, where $\mathcal{V}$ includes different types of nodes such as food items, user profile, health conditions and nutrition tags. In this paper, we further extend query node $\mathcal{V}_Q$ and agent nodes $\mathcal{V}_A$ into the knowledge graph. Specifically, agent nodes $\mathcal{V}_A$ represent candidate agents, each defined by its prompts and strategy description, while query nodes $v_q \in \mathcal{V}_Q$ are initialized by a contextual encoder. As such, all nodes are embedded into the shared textual space to make message passing meaningful.

\begin{figure*}[t]
	\centering
	\includegraphics[width=1\linewidth]{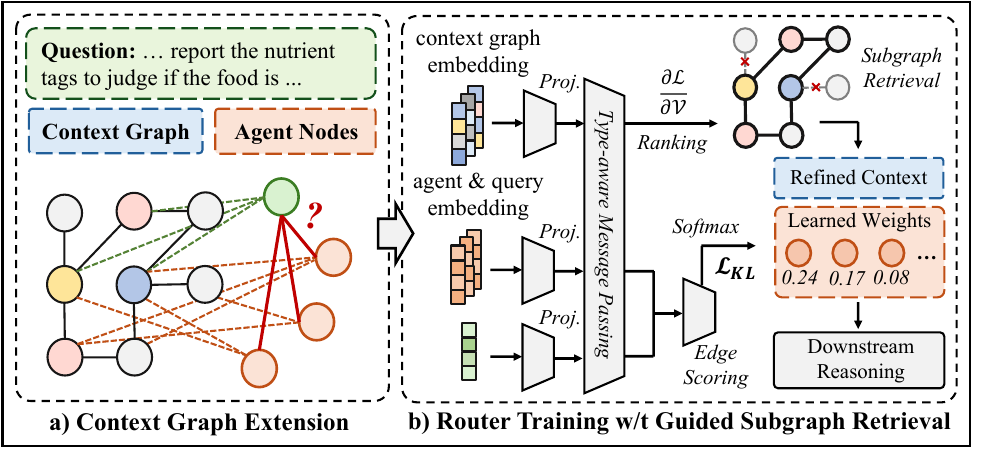}
        \vspace{-20pt}
	\caption{Overview of our proposed framework. (a) shows the KG extension process, where QA instances are extended into context graphs with query and agent nodes linked to nutritional entities. (b) shows our type-aware GNN router, which propagates contextual signals and guides gradient-based subgraph retrieval, refining the graph and producing agent importance weights for downstream collaboration and reasoning.}
        \vspace{-15pt}
    \label{fig:framework}
\end{figure*}

To further capture both the static semantic structure of the domain and the dynamic preferences of agent routing, we define: 1) Query–entity edges, which connect each query to the entities it explicitly mentions, thereby grounding the question within its evidential context. 2) Agent–entity edges, which capture the perspectives of different agents. By prompting each agent to identify the entities it finds most relevant, we encode how distinct agents selectively attend to contextual information. 3) Query–agent edges, which are not fixed but remain \textit{trainable}. These edges carry the routing signal that the model learns to optimize; their existence and weights determine which agents are activated for collaboration and how their contributions are combined. In this formulation (Figure~\ref{fig:framework}), the frozen edges ensure contextual grounding, while the adaptive query–agent edges provide a mechanism for dynamic and task-specific routing.

\subsection{Problem Formulation}
Formally, we study the problem of designing an optimized LLM-based agent collaboration scheme for a fixed graph QA task. Let $\mathcal{A}=\{a_1,\ldots,a_n\}$ denote the pool of available agents (each agent defined by a backbone LLM and an interaction strategy/prompting style), $\mathcal{X}$ the input/query space, $\mathcal{G}$ the context graphs of the queries, and $\mathcal{Y}$ the output space. For $x\in\mathcal{X}$, each agent $a\in\mathcal{A}$ produces a candidate $y_a(x, \mathcal{G})\in\mathcal{Y}$. Our overarching goal is to learn an optimal weighted combination of agents in $\mathcal{A}$ that maximizes task-level performance in the fixed downstream setting. Specifically, within this KG, the problem reduces to learning a function $f_\theta$ that scores query–agent pairs by propagating signals along graph edges:
\[
s(q,a) = f_\theta(q,a;\mathcal{G}),
\]
where $s(q,a)$ estimates the utility of including agent $a$ when solving query $q$ for the given task and $\mathcal{G}$ here represents the corresponding context graph. The model then computes a weighted combination of agents:
\[
\hat{y}(q) = \phi\!\left(\{\,y_a(q),\,w_a(q)\,:\, a \in \mathcal{A}\}\right),
\]
with weights $w_a(q) \propto \exp(s(q,a))$, and $\phi$ denoting an aggregation rule such as voting, reranking, or learned fusion. Framing the problem in this way refines the high-level goal of "finding the best collaboration scheme" into the concrete task of learning graph-supervised scores for query–agent pairs, which in turn yield optimized weightings of agents for a fixed downstream task. 

It is worth noting our formulation builds on the assumption that no single agent or backbone uniformly dominates; rather, their strengths and weaknesses are task- and backbone-dependent. This premise is supported by extensive prior works \cite{chen2024agentverse, chen2024reconcile} and later in our experiments, which consistently shows that different LLMs or prompting strategies excel in different scenarios.

\section{Methodology}

\subsection{Agent Collaboration Training}
Given the constructed knowledge graph, the central challenge is to train a router that can decide which agents are most valuable for solving a given query. Prior approaches often depend on manually crafted rules or the use of LLMs as external judges. Such heuristics lack adaptability and fail to capture complex contextual signals. In contrast, we frame routing as a supervised learning problem, allowing the model to discover fine-grained dependencies among queries, entities, and agents, rather than relying on rigid voting schemes.

Our solution employs a heterogeneous Graph Neural Network designed for type-sensitive message passing. Each node is first mapped into a unified latent representation through a type-specific projection $\mathrm{Proj}_{\tau(v)}$. For an edge $(u \xrightarrow{\psi} v)$ of relation type $\psi$, the propagated message is computed as
\[
m^{(l,\psi)}_{u \to v} = \mathrm{Proj.}\!\left(W^{(l)}_{\psi}\, h^{(l-1)}_u\right),
\]
and neighbor messages of the same type are aggregated by averaging. The contributions of different edge types are then merged with learnable gates, yielding the update
\[
h^{(l)}_v = U^{(l)}_{\tau(v)}\!\Big(h^{(l-1)}_v \,\|\, \sum_{\psi \in \Psi(v)} w^{(l)}_\psi \cdot \tilde{m}^{(l,\psi)}_v \Big),
\]
where $\tau(v)$ indicates node type, $w^{(l)}_\psi$ is a trainable coefficient per edge type, $U^{(l)}_{\tau(v)}$ is a type-dependent update operator, and $\|$ denotes concatenation. This mechanism enables each node to refine its embedding by combining its prior state with heterogeneous relational evidence.

After $L$ layers, the query representation $h^{(L)}_q$ incorporates contextual information, while agent embeddings $h^{(L)}_a$ encode their relative competence for the query. A routing score is then assigned by
\[
s(q,a) = \mathrm{MLP}\!\left(h^{(L)}_q \,\|\, h^{(L)}_a\right),
\]
which is normalized over all agents with a softmax:
\[
p_\theta(a \mid q,\mathcal{G}) = \mathrm{softmax}_{a \in \mathcal{A}} \big(s(q,a)\big).
\]

Supervision is provided by empirical agent performance. For each query, agents are evaluated and their F1 scores converted into a target distribution $p^*(a \mid q)$ using a temperature-scaled softmax. This produces smoother guidance than one-hot labels, emphasizing not only the best agent but also secondary contributors. Training minimizes the KL divergence
\[
\mathcal{L}_{\mathrm{KL}}(q) = \sum_{a \in \mathcal{A}} p^*(a \mid q)\, \log \frac{p^*(a \mid q)}{p_\theta(a \mid q,\mathcal{G})}.
\]

KL divergence is particularly well-suited here: unlike cross-entropy, which concentrates solely on the top class, or mean squared error, which poorly models probability distributions, KL captures the full relative structure among agents and maintains informative gradients. This encourages the router to learn balanced allocation across complementary agents, stabilizing training and promoting collaborative diversity.

During inference, the router outputs the distribution $p_\theta(a \!\mid\! q,\mathcal{G})$ over agents. Final predictions are assembled by weighted ensembling:
\[
\hat{y}(q) = \phi\!\left(\{\, y_a(q),\, p_\theta(a \mid q,\mathcal{G}) : a \in \mathcal{A}\}\right),
\]
where $y_a(q)$ denotes agent $a$’s answer and $\phi$ is a weighted aggregation rule such as probabilistic voting. Thus, the learned distribution directly determines how much influence each agent exerts in producing the final answer, yielding a principled, context-aware collaboration strategy.

\begin{table*}[t]
\centering
\resizebox{\textwidth}{!}{
\begin{tabular}{c *{9}{c}}
\toprule
\multirow{2}{*}{\textbf{Method}} & 
\multicolumn{3}{c}{\textbf{Sparse}} & 
\multicolumn{3}{c}{\textbf{Standard}} & 
\multicolumn{3}{c}{\textbf{Complex}} \\
\cmidrule(lr){2-4} \cmidrule(lr){5-7} \cmidrule(lr){8-10}
& Accuracy & Precision & F1 & Accuracy & Precision & F1 & Accuracy & Precision & F1 \\
\midrule
KAPING          & 17.53 & 20.75 & 33.94 & 45.93 & 46.24 & 62.72 & 68.83 & 71.29 & 80.93 \\
ToG             & 24.39 & 29.86 & 43.33 & 61.89 & 67.93 & 74.64 & 61.53 & 81.19 & 73.03 \\
\midrule
Raw             & \underline{29.34$\pm$0.74} & 32.51$\pm$0.59 & 49.41$\pm$0.91 & 
                   64.11$\pm$0.31 & \underline{79.29$\pm$0.47} & 82.40$\pm$0.42 &
                   72.03$\pm$0.23 & 72.96$\pm$0.24 & 81.86$\pm$0.18 \\
CoT             & 29.21$\pm$1.43 & \underline{32.54$\pm$1.12} & 48.10$\pm$1.49 &
                   62.91$\pm$0.86 & 76.51$\pm$0.97 & 80.27$\pm$0.64 &
                   72.40$\pm$0.16 & 72.93$\pm$0.16 & 82.05$\pm$0.18 \\
MAD             & 27.10$\pm$1.32 & 30.59$\pm$1.24 & 46.39$\pm$1.14 &
                   61.73$\pm$0.54 & 76.05$\pm$0.50 & 80.34$\pm$0.34 &
                   71.80$\pm$0.27 & 72.77$\pm$0.36 & 81.88$\pm$0.27 \\
React-Reflect  & 25.14$\pm$1.39 & 28.81$\pm$1.37 & 47.72$\pm$0.53 &
                   66.67$\pm$0.59 & 82.36$\pm$0.56 & 84.89$\pm$0.27 &
                   71.92$\pm$0.34 & 72.80$\pm$0.03 & 81.73$\pm$0.19 \\
SC              & 25.71$\pm$0.29 & 29.64$\pm$0.38 & 45.52$\pm$0.10 &
                   68.76$\pm$0.20 & 85.81$\pm$0.32 & 86.36$\pm$0.19 &
                   72.75$\pm$0.39 & 73.43$\pm$0.39 & 82.45$\pm$0.28 \\
Summary         & 27.45$\pm$0.77 & 31.23$\pm$0.76 & 47.12$\pm$1.25 &
                   66.72$\pm$0.19 & 84.00$\pm$0.31 & 85.45$\pm$0.17 &
                   71.28$\pm$0.65 & 72.56$\pm$0.35 & 81.60$\pm$0.36 \\
\midrule
Majority Vote   & 22.61$\pm$0.13 & 25.58$\pm$0.17 & 44.59$\pm$0.28 &
                   57.35$\pm$0.43 & 61.79$\pm$0.33 & 75.26$\pm$0.20 &
                   52.43$\pm$0.77 & 77.50$\pm$0.61 & 71.73$\pm$0.29 \\
HybridLLM      & 28.02$\pm$0.59 & 30.91$\pm$0.42 & 47.76$\pm$0.74 &
                   34.45$\pm$0.12 & 34.45$\pm$0.12 & 49.25$\pm$0.22 &
                   72.75$\pm$0.29 & 73.19$\pm$0.14 & 82.31$\pm$0.21 \\
LLM-Blender     & 21.20$\pm$0.98 & 24.26$\pm$0.95 & 46.73$\pm$1.43 &
                   45.72$\pm$0.55 & 50.90$\pm$0.32 & 68.32$\pm$0.83 &
                   51.79$\pm$0.17 & 72.02$\pm$0.21 & 75.50$\pm$0.69 \\
\midrule
NG-Router w/o SR     
                 & 29.11$\pm$0.32 & 32.24$\pm$0.40 & \underline{49.71$\pm$0.65} &
                   \underline{76.33$\pm$1.17} & 78.21$\pm$1.01 & \underline{89.14$\pm$0.59} &
                   \underline{74.63$\pm$0.19} & \underline{77.58$\pm$0.12} & \underline{85.95$\pm$0.06} \\
NG-Router w/ SR 
                 & \textbf{57.49$\pm$0.25} & \textbf{61.58$\pm$0.28} & \textbf{75.29$\pm$0.17} &
                   \textbf{94.10$\pm$2.58} & \textbf{97.65$\pm$1.71} & \textbf{97.73$\pm$1.24} &
                   \textbf{88.19$\pm$1.92} & \textbf{95.04$\pm$0.18} & \textbf{92.53$\pm$1.24} \\
\bottomrule
\end{tabular}
}
\vspace{-5pt}
\caption{Performance comparison across Sparse, Standard, and Complex settings. We report both the version with and without the subgraph retrieval (SR). Additionally, we report the mean and standard deviation for all results for three runs. Best results are in \textbf{bold}, second best are \underline{underlined}.}
\label{tab:benchmark_results}
\vspace{-5pt}
\end{table*}

\subsection{Graph-Supervised Subgraph Retrieval}

A critical challenge in nutritional QA is the excessive volume of contextual information, where presenting the full graph to the router dilutes reasoning and introduces noise. To address this, we further propose a subgraph retrieval module that leverages training signals to identify and retain the most salient entities for each query.

Let $\mathcal{V}_E$ denote entity nodes and $\mathbf{h}_i$ the last layer of the embeddings of entity $v_i\in\mathcal{V}_E$. For a given query $q$, we obtain the agent scores $s(q,a)$ and routing distribution $p_\theta(a \mid q,\mathcal{G})$ as defined in the previous subsection. To measure the contribution of each entity, we compute gradients of the routing objective with respect to $\mathbf{h}_i$. Specifically, we define the salience of entity $v_i$ as
\[
\alpha_i \;=\; \Big\|\, \nabla_{\mathbf{h}_i} \, \mathcal{L}_{\mathrm{KL}}(q) \,\Big\|_2 ,
\]
which quantifies how strongly the training loss depends on $v_i$. Entities with higher $\alpha_i$ exert greater influence on agent routing and are therefore more informative. At inference, we evaluate $\alpha_i$ for all entities in $\mathcal{V}_E$ associated with $q$ and filter out the entities that are not important enough, in our case the threshold $\tau$ is 0.01:
\[
\mathcal{V}^{*} \;=\; \mathcal{f_\theta}\!\left(\{(v_i,\alpha_i): v_i \in \mathcal{V}_E\},\, \theta > \tau\right).
\]
We then construct the induced subgraph $\mathcal{G}^*$ over $\mathcal{V}^*$, preserving their original edges. The downstream reasoning at inference time will use $\mathcal{G}^*$ instead of $\mathcal{G}$ as the context graph, therefore $y_a(x, \mathcal{G^*})$. This ensures that downstream reasoning attends to the most critical contextual evidence while filtering out redundant information. The final output will be the weighted vote result of the full agent set $\mathcal{A}$.

\begin{table}[t]
\centering
\resizebox{\linewidth}{!}{
\begin{tabular}{l l ccc}
\toprule
\textbf{Type} & \textbf{Metric} & \textbf{Sparse} & \textbf{Standard} & \textbf{Complex} \\
\midrule
\multirow{4}{*}{Node} 
  & Original      & 26.60 & 27.85 & 30.89 \\
  & Retrieval     & 7.87  &  9.71 & 13.67 \\
  & Drop (\%)     & 70.41 & 65.13 & 55.75 \\
\midrule
\multirow{3}{*}{Edge} 
  & Original      & 51.09 & 55.56 & 67.60 \\
  & Retrieval     & 10.78 & 19.95 & 30.24 \\
  & Drop (\%)     & 78.90 & 64.09 & 55.27 \\
\midrule
\multirow{3}{*}{SNR} 
  & Original      & 16.40 & 27.70 & 31.60 \\
  & Retrieval     & 50.83 & 70.44 & 70.80 \\
  & Raise (\%)     & 209.94 & 154.30 & 124.05 \\
\bottomrule
\end{tabular}
}
\vspace{-5pt}
\caption{Graph size reduction before and after applying subgraph retrieval. Node and edge statistics are shown for Sparse, Standard, and Complex settings. Signals-to-Noise Ratio (SNR) indicates the proportion of useful nodes in the context graph.}
\vspace{-15pt}
\label{tab:rag_shrinkage}
\end{table}

\begin{figure}[t]
	\centering
	\includegraphics[width=1\linewidth]{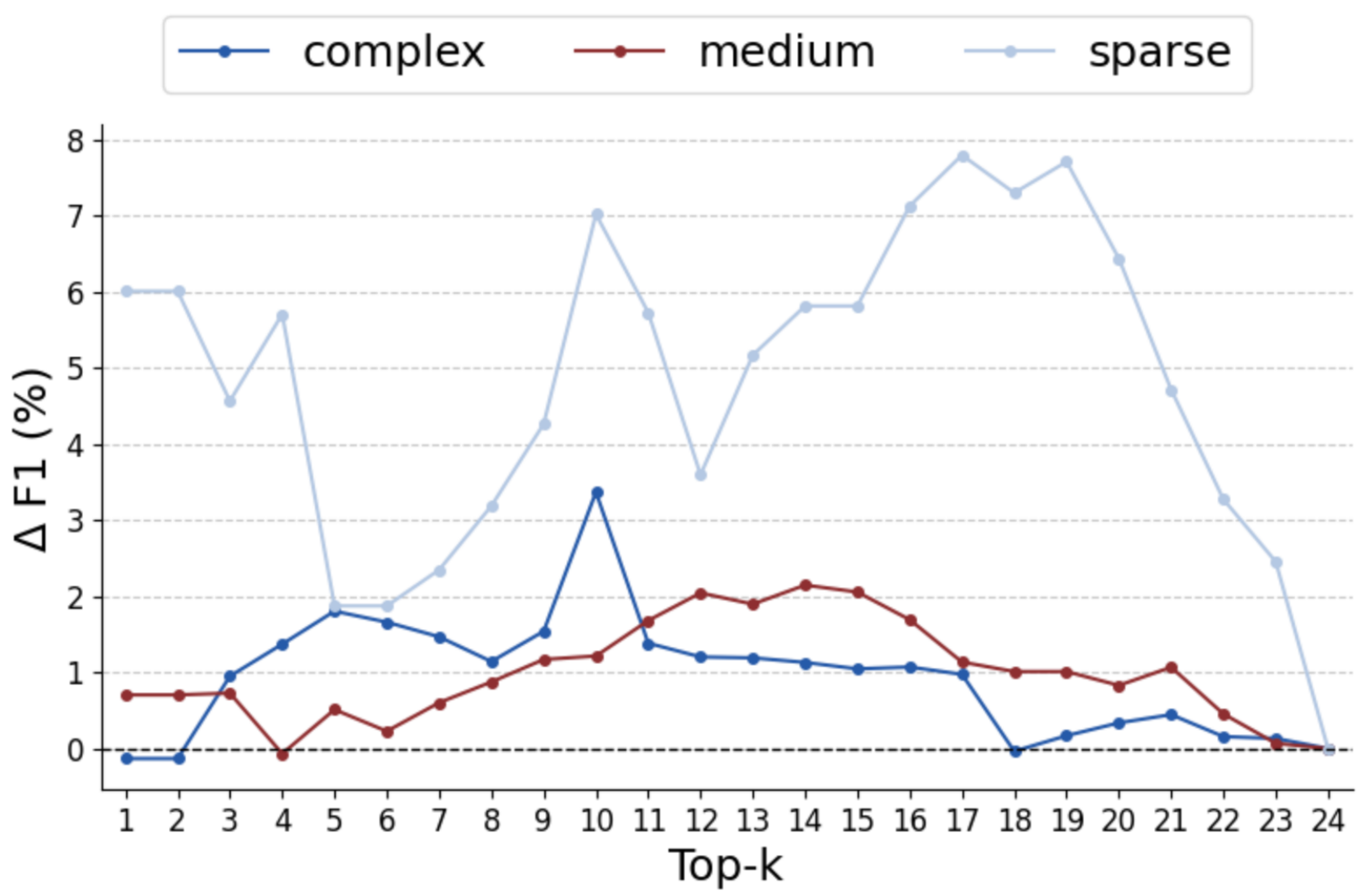}
        \vspace{-20pt}
	\caption{Percentage change ($\Delta$) in F1 relative to k=24, used as the base (0\%). Curves show how performance varies with the top $k$ agent clipped across datasets.}
        \vspace{-15pt}
    \label{fig:topk}
\end{figure}

\begin{table*}[ht!]
\centering
\resizebox{\textwidth}{!}{
\begin{tabular}{llccccccc}
\toprule
\textbf{Setting} & \textbf{Backbone} & \textbf{CoT} & \textbf{MAD} & \textbf{Raw} & \textbf{ReAct-Reflect} & \textbf{SC} & \textbf{Summary} & \textbf{Best Agent} \\
\midrule
\multirow{4}{*}{Sparse} 
 & GPT-4o-mini & 48.10$\pm$1.49 & 46.39$\pm$1.14 & 49.41$\pm$0.91 & 47.09$\pm$1.39 & 45.52$\pm$0.10 & 47.12$\pm$1.25 & 78.40 \\
 & Llama-3.2-3B & 45.87$\pm$0.68 & 45.33$\pm$0.72 & 46.46$\pm$0.93 & 47.72$\pm$0.53 & 45.38$\pm$0.13 & 46.33$\pm$0.07 & 56.90 \\
 & Mistral-7B-Instruct & 46.73$\pm$1.43 & 44.70$\pm$0.43 & 47.29$\pm$0.27 & 44.23$\pm$0.88 & 42.42$\pm$0.50 & 47.10$\pm$0.53 & 54.38 \\
 & Qwen2.5-7B-Instruct & 43.03$\pm$0.14 & 40.42$\pm$0.33 & 42.86$\pm$0.21 & 42.52$\pm$0.03 & 42.29$\pm$0.03 & 42.20$\pm$0.32 & 81.70 \\
\midrule
\multirow{4}{*}{Standard} 
 & GPT-4o-mini & 52.25$\pm$3.79 & 58.54$\pm$6.41 & 52.11$\pm$4.47 & 53.02$\pm$2.71 & 52.35$\pm$4.61 & 54.67$\pm$3.85 & 51.09 \\
 & Llama-3.2-3B & 64.02$\pm$0.78 & 70.31$\pm$0.16 & 64.78$\pm$1.65 & 71.66$\pm$1.18 & 66.30$\pm$1.02 & 72.65$\pm$1.31 & 74.09 \\
 & Mistral-7B-Instruct & 68.32$\pm$0.83 & 79.40$\pm$0.76 & 66.97$\pm$0.21 & 73.95$\pm$1.22 & 67.85$\pm$0.73 & 76.98$\pm$0.78 & 79.71 \\
 & Qwen2.5-7B-Instruct & 80.27$\pm$0.64 & 80.34$\pm$0.34 & 82.40$\pm$0.42 & 84.89$\pm$0.27 & 86.36$\pm$0.19 & 85.45$\pm$0.17 & 86.47 \\
\midrule
\multirow{4}{*}{Complex} 
 & GPT-4o-mini & 82.05$\pm$0.18 & 81.88$\pm$0.27 & 81.86$\pm$0.18 & 81.73$\pm$0.19 & 82.45$\pm$0.28 & 81.60$\pm$0.36 & 82.29 \\
 & Llama-3.2-3B & 66.97$\pm$1.21 & 59.79$\pm$0.37 & 64.97$\pm$0.04 & 61.54$\pm$0.12 & 62.01$\pm$0.33 & 59.23$\pm$0.58 & 67.67 \\
 & Mistral-7B-Instruct & 75.50$\pm$0.69 & 68.80$\pm$0.50 & 75.59$\pm$0.06 & 73.12$\pm$0.02 & 75.01$\pm$0.13 & 70.58$\pm$0.13 & 75.90 \\
 & Qwen2.5-7B-Instruct & 74.37$\pm$13.48 & 69.95$\pm$9.80 & 74.37$\pm$12.72 & 72.30$\pm$11.20 & 71.13$\pm$10.68 & 71.59$\pm$11.71 & 82.15 \\
\bottomrule
\end{tabular}
}
\vspace{-5pt}
\caption{Performance of different agent designs across Sparse, Standard, and Complex settings on four LLM backbones. We report mean F1 with standard deviation. The last column shows the best-performing agent within each backbone.}
\vspace{-15pt}
\label{tab:f1_results}
\end{table*}

\section{Experiments}
\subsection{Experiment Setup}
\textbf{For benchmark,} we evaluate the label generation task on three datasets from the NGQA benchmark \cite{zhang2024ngqa}, namely \textit{sparse}, \textit{standard}, and \textit{complex}, designed for nutrition question reasoning. We follow the evaluation settings and evaluation metrics of the NGQA benchmark. The detailed description of the benchmark and dataset, as well as training settings, are provided in Appendix~\ref{app:benchmarks}.  

\noindent\textbf{For baselines,} we consider three categories: (1) the original baselines reported in NGQA, namely KAPING \cite{baek2023knowledge} and ToG \cite{sunthink}; (2) the best-performing single-agent and multi-agent designs across several LLM backbones. Specifically, we include Raw (the basic LLM method), Chain-of-Thought (CoT) \cite{wei2022chain}, Self-Consistency (SC) \cite{wang2023selfconsistency}, ReAct-Reflection \cite{yao2023react, shinn2023reflexion}, Multi-Agent Debate (MAD) \cite{du2024mad}, and Multi-Agent Summary. Our method also draws from these six agent designs as its candidate pool; (3) Strong ensembling method such as Majority Vote and the state-of-the-art LLM routing baselines, including LLM-Blender \cite{jiang2023llm} and HybridLLM \cite{ding2024hybridllm}. For experiments, we use four widely adopted LLM backbones of comparable scale but from different providers: Llama-3.2-3B-Instruct \cite{meta2024llama3_2_3b_instruct}, Qwen2.5-7B-Instruct-Turbo \cite{qwen2025qwen2_5_7binstruct}, Mistral-7B-Instruct-v0.2 \cite{mistral2024_mistral7b_instruct_v0_2}, and GPT-4o-mini \cite{openai_gpt4o_mini}. All baselines are evaluated under the same settings, and results are averaged over three consecutive runs to mitigate randomness. Additional implementation details of the baselines are provided in Appendix~\ref{app:baselines}.

\subsection{Main Result}

We present the main results of NG-Router against baseline methods in Table~\ref{tab:benchmark_results}. Across all three benchmark settings, our approach consistently achieves superior performance, demonstrating the effectiveness of the proposed routing framework. Beyond the overall gains, several noteworthy insights emerge. First, the integration of subgraph retrieval yields substantial improvements. In contrast, traditional rule-based retrieval methods such as KAPING and ToG perform markedly worse. This highlights the advantage of our approach: by leveraging message-passing signals, the retrieval process can identify informative nodes even at longer distances, whereas heuristic methods tend to overemphasize immediate neighbors and thus overlook critical evidence. Second, the impact of subgraph retrieval varies across benchmarks. On the sparse setting, performance improves by more than 50\%, whereas on the complex dataset the gains are closer to 10\%. Table~\ref{tab:rag_shrinkage} sheds light on this phenomenon: the retrieval process prunes over 70\% of edges and nodes, effectively filtering noise in the sparse setting where relevant information is scarce. By contrast, in the complex setting, where signal-to-noise (SNR) ratios (Table~\ref{tab:snr_stats} in Appendix~\ref{app:benchmarks}) are inherently higher, the marginal benefit of pruning is smaller. We can observe that the retrieved graph almost doubled in SNR, which also indicates the raise in the performance. Finally, we also examine the best single-agent designs across different LLM backbones. Results reveal that agent effectiveness varies by task, underscoring the necessity of learning adaptive collaboration schemes rather than relying on fixed heuristics. By coordinating agents through a weighted-voting mechanism learned via routing, NG-Router consistently outperforms all individual agents, illustrating the benefit of principled collaboration in multi-agent reasoning.

\subsection{Top K Pruning}
During our experiment, we report the full routing results, however, we notice that pruning the long tail of low-quality or noisy agents is helpful to reduce variance and sharpens the aggregation of useful reasoning patterns. This suggests that agent routing benefits not only from diversity but also from judicious selection, where a smaller yet high-quality subset provides a better balance between complementarity and noise. As can be seen in Figure~\ref{fig:topk}, full agent routing is usually not the optimal solution, whereas pruning the agent number to 10-15 would yield better overall performance.

\begin{table*}[t]
\centering
\resizebox{\textwidth}{!}{
\begin{tabular}{lccccccccc}
\toprule
\multirow{2}{*}{\textbf{Method}} & 
\multicolumn{4}{c}{\textbf{Binary Classification}} & 
\multicolumn{5}{c}{\textbf{Natural Text Generation}} \\
\cmidrule(lr){2-5} \cmidrule(lr){6-10}
& Accuracy & Recall & Precision & F1 & ROUGE-1 & ROUGE-2 & ROUGE-L & BLEU & BERT \\
\midrule
Majority Vote & 53.38$\pm$0.00 & 71.00$\pm$0.00 & 68.27$\pm$0.00 & 69.67$\pm$0.00 & 60.32$\pm$0.10 & 60.84$\pm$0.66 & 60.32$\pm$0.10 & 41.73$\pm$1.85 & 86.29$\pm$0.02 \\
Hybrid LLM    & 44.23$\pm$0.61 & 61.33$\pm$0.58 & 61.33$\pm$0.58 & 61.33$\pm$0.58 & 69.55$\pm$0.70 & 62.65$\pm$1.05 & 68.65$\pm$0.74 & 46.86$\pm$1.00 & 95.81$\pm$0.09 \\
LLM Blender   & 50.38$\pm$0.00 & 67.00$\pm$0.00 & 67.00$\pm$0.00 & 67.00$\pm$0.00 & 68.17$\pm$0.51 & 63.26$\pm$0.64 & 67.32$\pm$0.46 & 49.85$\pm$0.77 & 95.74$\pm$0.04 \\
NG-Router     & 55.06$\pm$2.10 & 71.00$\pm$1.73 & 71.00$\pm$1.73 & 71.00$\pm$1.73 & 70.88$\pm$0.17 & 65.51$\pm$0.37 & 69.73$\pm$0.06 & 51.55$\pm$0.25 & 96.14$\pm$0.00 \\
\bottomrule
\end{tabular}
}
\vspace{-5pt}
\caption{Comparison of \textbf{binary classification metrics} (Accuracy, Recall, Precision, F1) and \textbf{natural text generation metrics} (ROUGE, BLEU, BERT). All values are reported with mean $\pm$ standard deviation.}
\vspace{-15pt}
\label{tab:transfer}
\end{table*}

\subsection{Transfer Analysis}

Beyond the label generation task, we further evaluate the transferability of our approach by applying the trained multi-label generation model to the two additional tasks defined in the NGQA benchmark: binary classification and natural text generation. As shown in Table~\ref{tab:transfer}, NG-Router achieves the strongest overall performance across both task types. For binary classification, our method yields improvements in both precision and F1 over all zero-shot baselines, demonstrating that the learned representations are sufficiently robust to support accurate decision-making when the task is simplified to binary outcomes. For natural text generation, NG-Router consistently surpasses competing methods on ROUGE, BLEU, and BERT metrics, indicating that the routing framework not only preserves semantic fidelity but also enhances fluency and informativeness in generated answers. These results confirm that the knowledge-graph–guided design of NG-Router provides a transferable advantage, enabling effective adaptation to multiple downstream tasks. Taken together, this analysis underscores the generalizability of our framework and its potential to serve as a versatile foundation for reasoning in diverse nutritional QA scenarios.

\subsection{Hyperparameter Analysis}

\begin{figure}[t]
	\centering
	\includegraphics[width=1\linewidth]{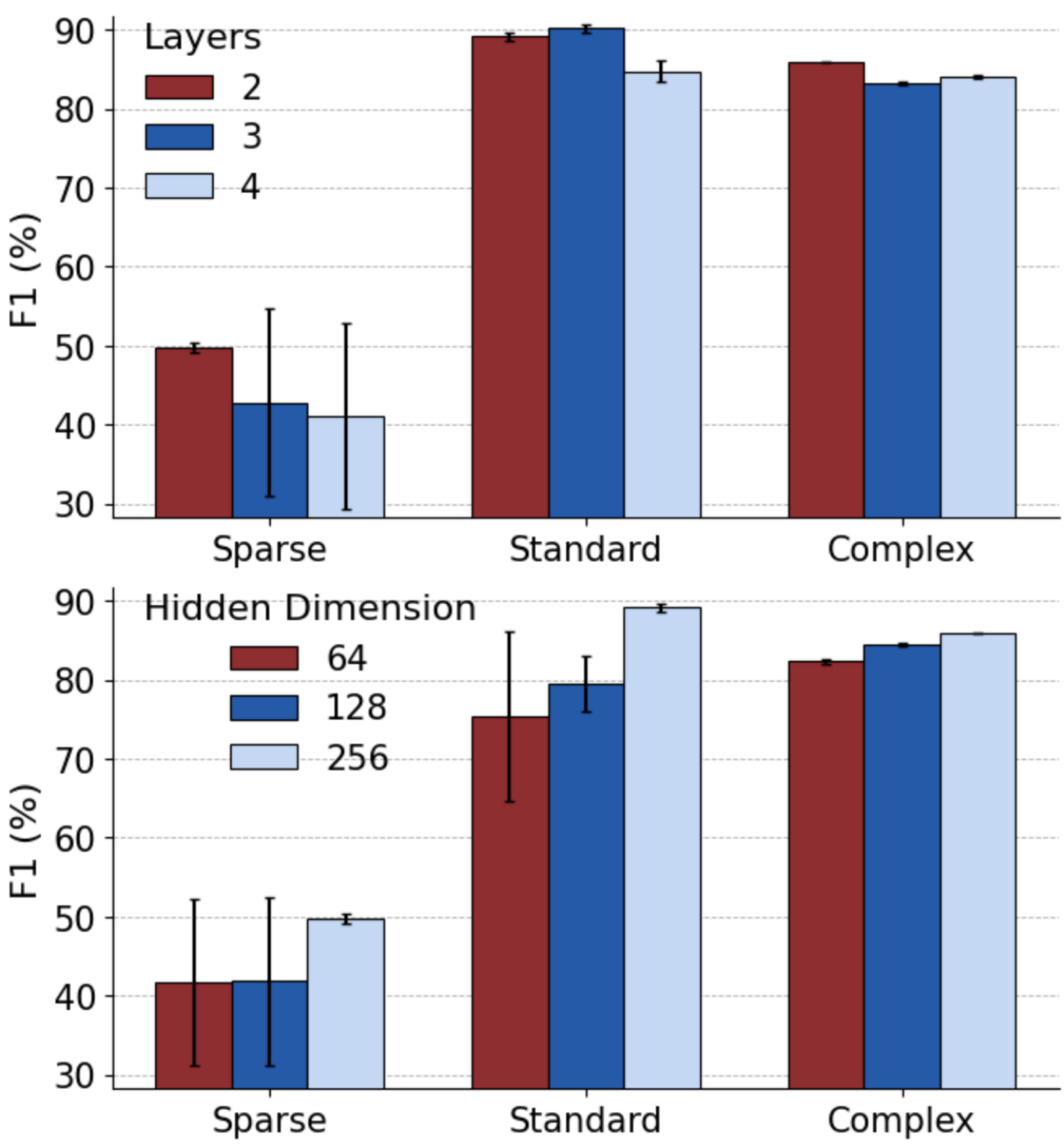}
        \vspace{-20pt}
	\caption{F1 performance on the three datasets, varying Layers (top) and Hidden Dimensions (bottom). Error bars denote standard deviations.}
        \vspace{-15pt}
    \label{fig:hyper}
\end{figure}

In this section, we study the effect of architectural hyper-parameters. Figure~\ref{fig:hyper} presents the results of varying the number of layers and hidden dimensions. Across all benchmarks, we find that performance differences between configurations are moderate yet systematic. Increasing the number of layers does not consistently improve F1: while three layers slightly outperform two on the \textit{Standard} dataset, deeper settings yield lower scores on \textit{Sparse} and \textit{Complex}, with larger variances observed for deeper models. This suggests diminishing returns and instability with depth, especially in multi-hop reasoning tasks. By contrast, enlarging the hidden dimension provides stable gains across benchmarks. 
Performance improves monotonically from 64 to 256 dimensions, with reduced variance at higher capacities, indicating that representational richness plays a more robust role than depth in this setting. Overall, these findings suggest that a shallow architecture with a larger hidden size strikes a better trade-off between accuracy and stability, highlighting the importance of capacity over depth in designing models for diverse QA benchmarks. Additionally, we report the original agent performance used to calculate the best agent scores, as can be seen in Table~\ref{tab:f1_results}.

\section{Conclusion}
We proposed \textsc{NG-Router}, a graph-supervised framework for adaptive multi-agent collaboration in nutritional QA, an important domain specific field. By extending context graphs with query and agent nodes, our framework learns task-specific routing weight distributions to search for optimal agent collaboration schemes, while gradient-guided subgraph retrieval prunes irrelevant entities to reduce contextual overload. Beyond its empirical gains, \textsc{NG-Router} offers practitioners a principled way to harness the complementary strengths of heterogeneous agents without relying on costly trial-and-error or heuristic ensembling. This makes it a robust and generalizable tool for deploying reliable multi-agent systems in complex domain specific reasoning tasks.

\newpage
\section*{Limitations}
This work is limited by its reliance on the NGQA benchmark, which is derived from U.S.-centric dietary surveys. While this scope may restrict direct generalizability to other populations, domains, or languages, NGQA remains the most comprehensive benchmark currently available for personalized nutrition reasoning. Our model is intentionally designed for this domain-specific setting, where carefully curated knowledge graphs and diverse dietary-health annotations provide a rigorous testbed. We view this as an essential first step, and leave the extension to broader public benchmarks and multilingual dietary datasets to future work.

Our evaluation also relies on automatic metrics such as accuracy, precision, and F1. These provide standardized comparisons with baselines and highlight reasoning quality, but they do not capture clinical or behavioral outcomes. Similarly, the design of the knowledge graph and the gradient-based subgraph retrieval involve thresholding choices that may introduce bias or limit stability across different datasets. We adopt these abstractions to ensure scalability and systematic experimentation, and envision future refinements that incorporate adaptive graph schemas, richer outcome measures, and broader evaluation settings.

\newpage
\bibliography{reference}

\begin{thebibliography}{77}
\providecommand{\natexlab}[1]{#1}

\bibitem[{Afshin et~al.(2019)Afshin, Sur, Fay, Cornaby, Ferrara, Salama, and Murray}]{Afshin_2019_DietaryRisks}
Ashkan Afshin, Patrick~J Sur, Kairsten~A Fay, Leslie Cornaby, Giannina Ferrara, Jason~S Salama, and Christopher J~L Murray. 2019.
\newblock Health effects of dietary risks in 195 countries, 1990–2017: a systematic analysis for the global burden of disease study 2017.
\newblock \emph{The Lancet}.

\bibitem[{Alibaba(2025)}]{qwen2025qwen2_5_7binstruct}
Qwen Alibaba. 2025.
\newblock Qwen2.5-7b-instruct.
\newblock \url{https://huggingface.co/Qwen/Qwen2.5-7B-Instruct}.

\bibitem[{Baek et~al.(2023)Baek, Aji, and Saffari}]{baek2023knowledge}
Jinheon Baek, Alham~Fikri Aji, and Amir Saffari. 2023.
\newblock Knowledge-augmented language model prompting for zero-shot knowledge graph question answering.
\newblock In \emph{ACL}.

\bibitem[{B{\"o}lz et~al.(2023)B{\"o}lz, Nurbakova, Calabretto, Gerl, Brunie, and Kosch}]{bolz2023hummus}
Felix B{\"o}lz, Diana Nurbakova, Sylvie Calabretto, Armin Gerl, Lionel Brunie, and Harald Kosch. 2023.
\newblock Hummus: A linked, healthiness-aware, user-centered and argument-enabling recipe data set for recommendation.
\newblock In \emph{RecSys}.

\bibitem[{Bondevik et~al.(2024)Bondevik, Bennin, Babur, and Ersch}]{bondevik2024systematic}
Jon~Nicolas Bondevik, Kwabena~Ebo Bennin, {\"O}nder Babur, and Carsten Ersch. 2024.
\newblock A systematic review on food recommender systems.
\newblock \emph{Expert Systems with Applications}.

\bibitem[{Cao et~al.(2023)Cao, Xu, Yang, Wang, Zhang, Wang, Chen, and Yang}]{cao2023pre}
Yuxuan Cao, Jiarong Xu, Carl Yang, Jiaan Wang, Yunchao Zhang, Chunping Wang, Lei Chen, and Yang Yang. 2023.
\newblock When to pre-train graph neural networks? from data generation perspective!
\newblock In \emph{KDD}.

\bibitem[{CDC(2020)}]{CDC_2020_Obesity}
CDC. 2020.
\newblock \href {https://www.cdc.gov/obesity/data/adult.html} {Adult obesity facts}.

\bibitem[{Chen et~al.(2024{\natexlab{a}})Chen, Saha, and Bansal}]{chen2024reconcile}
Justin Chih-Yao Chen, Swarnadeep Saha, and Mohit Bansal. 2024{\natexlab{a}}.
\newblock Reconcile: Round-table conference improves reasoning via consensus among diverse {LLM}s.
\newblock In \emph{ACL}.

\bibitem[{Chen et~al.(2024{\natexlab{b}})Chen, Jiang, Lin, Kwok, and Zhang}]{chen2024routerdc}
Shuhao Chen, Weisen Jiang, Baijiong Lin, James Kwok, and Yu~Zhang. 2024{\natexlab{b}}.
\newblock Routerdc: Query-based router by dual contrastive learning for assembling large language models.
\newblock \emph{NeurIPS}.

\bibitem[{Chen et~al.(2024{\natexlab{c}})Chen, Su, Zuo, Yang, Yuan, Chan, Yu, Lu, Hung, Qian et~al.}]{chen2024agentverse}
Weize Chen, Yusheng Su, Jingwei Zuo, Cheng Yang, Chenfei Yuan, Chi-Min Chan, Heyang Yu, Yaxi Lu, Yi-Hsin Hung, Chen Qian, et~al. 2024{\natexlab{c}}.
\newblock Agentverse: Facilitating multi-agent collaboration and exploring emergent behaviors.
\newblock In \emph{ICLR}.

\bibitem[{Ding et~al.(2025)Ding, Mallick, Zhang, Wang et~al.}]{ding2025best}
D.~Ding, Ankur Mallick, Shaokun Zhang, Chi Wang, et~al. 2025.
\newblock Best-route: Adaptive llm routing with test-time optimal compute.
\newblock In \emph{ICML}.

\bibitem[{Ding et~al.(2024)Ding, Mallick, Wang, Sim, Mukherjee, Ruhle, Lakshmanan, and Awadallah}]{ding2024hybridllm}
Dujian Ding, Ankur Mallick, Chi Wang, Robert Sim, Subhabrata Mukherjee, Victor Ruhle, Laks~VS Lakshmanan, and Ahmed~Hassan Awadallah. 2024.
\newblock Hybridllm: Cost-efficient and quality-aware query routing.
\newblock In \emph{ICLR}.

\bibitem[{Du et~al.(2024)Du, Li, Torralba, Tenenbaum, and Mordatch}]{du2024mad}
Yilun Du, Shuang Li, Antonio Torralba, Joshua~B. Tenenbaum, and Igor Mordatch. 2024.
\newblock Improving factuality and reasoning in language models through multi-agent debate.
\newblock In \emph{ICML}.

\bibitem[{Fatemi et~al.(2023)Fatemi, Halcrow, and Perozzi}]{fatemi2023talk}
Bahare Fatemi, Jonathan Halcrow, and Bryan Perozzi. 2023.
\newblock Talk like a graph: Encoding graphs for large language models.
\newblock \emph{arXiv}.

\bibitem[{Feng et~al.(2025)Feng, Zhao, Liu, Yang, and Zhao}]{feng2024graphrouter}
Yihan Feng, Tianyu Zhao, Haotian Liu, Diyi Yang, and Tuo Zhao. 2025.
\newblock Graphrouter: Learning graph-based routing for large language model selection.
\newblock In \emph{ICLR}.

\bibitem[{Gao et~al.(2024)Gao, Qiao, Kan, Wen, He, and Li}]{gao2024two}
Yifu Gao, Linbo Qiao, Zhigang Kan, Zhihua Wen, Yongquan He, and Dongsheng Li. 2024.
\newblock Two-stage generative question answering on temporal knowledge graph using large language models.
\newblock \emph{arXiv}.

\bibitem[{Ge et~al.(2015)Ge, Ricci, and Massimo}]{ge2015health}
Mouzhi Ge, Francesco Ricci, and David Massimo. 2015.
\newblock Health-aware food recommender system.
\newblock In \emph{RecSys}.

\bibitem[{Guo et~al.(2024)Guo, Yang, Wang, Liu, Li, Tang, Li, and Wen}]{guo2024knowledgenavigator}
Tiezheng Guo, Qingwen Yang, Chen Wang, Yanyi Liu, Pan Li, Jiawei Tang, Dapeng Li, and Yingyou Wen. 2024.
\newblock Knowledgenavigator: Leveraging large language models for enhanced reasoning over knowledge graph.
\newblock \emph{Complex \& Intelligent Systems}.

\bibitem[{Hamilton et~al.(2017)Hamilton, Ying, and Leskovec}]{hamilton2017inductive}
Will Hamilton, Zhitao Ying, and Jure Leskovec. 2017.
\newblock Inductive representation learning on large graphs.
\newblock \emph{NeurIPS}.

\bibitem[{He et~al.(2024)He, Tian, Sun, Chawla, Laurent, LeCun, Bresson, and Hooi}]{he2024g}
Xiaoxin He, Yijun Tian, Yifei Sun, Nitesh~V Chawla, Thomas Laurent, Yann LeCun, Xavier Bresson, and Bryan Hooi. 2024.
\newblock G-retriever: Retrieval-augmented generation for textual graph understanding and question answering.
\newblock \emph{arXiv}.

\bibitem[{Hong et~al.(2024)Hong, Zhuge, Chen, Zheng, Cheng, Wang, Zhang, Wang, Yau, Lin et~al.}]{hongmetagpt}
Sirui Hong, Mingchen Zhuge, Jonathan Chen, Xiawu Zheng, Yuheng Cheng, Jinlin Wang, Ceyao Zhang, Zili Wang, Steven Ka~Shing Yau, Zijuan Lin, et~al. 2024.
\newblock Metagpt: Meta programming for a multi-agent collaborative framework.
\newblock In \emph{ICLR}.

\bibitem[{Jiang et~al.(2023{\natexlab{a}})Jiang, Ren, and Lin}]{jiang2023llm}
Dongfu Jiang, Xiang Ren, and Bill~Yuchen Lin. 2023{\natexlab{a}}.
\newblock {LLM}-blender: Ensembling large language models with pairwise ranking and generative fusion.
\newblock In \emph{ACL}.

\bibitem[{Jiang et~al.(2023{\natexlab{b}})Jiang, Zhou, Dong, Ye, Zhao, and Wen}]{jiang2023structgpt}
Jinhao Jiang, Kun Zhou, Zican Dong, Keming Ye, Wayne~Xin Zhao, and Ji-Rong Wen. 2023{\natexlab{b}}.
\newblock Structgpt: A general framework for large language model to reason over structured data.
\newblock In \emph{EMNLP}.

\bibitem[{Jiang et~al.(2024)Jiang, Zhou, Zhao, Song, Zhu, Zhu, and Wen}]{jiang2024kg}
Jinhao Jiang, Kun Zhou, Wayne~Xin Zhao, Yang Song, Chen Zhu, Hengshu Zhu, and Ji-Rong Wen. 2024.
\newblock Kg-agent: An efficient autonomous agent framework for complex reasoning over knowledge graph.
\newblock \emph{arXiv}.

\bibitem[{Jin et~al.(2024)Jin, Xie, Zhang, Roy, Zhang, Li, Li, Tang, Wang, Meng et~al.}]{jin2024graph}
Bowen Jin, Chulin Xie, Jiawei Zhang, Kashob~Kumar Roy, Yu~Zhang, Zheng Li, Ruirui Li, Xianfeng Tang, Suhang Wang, Yu~Meng, et~al. 2024.
\newblock Graph chain-of-thought: Augmenting large language models by reasoning on graphs.
\newblock In \emph{ACL}.

\bibitem[{Kim et~al.(2023)Kim, Kwon, Jo, and Choi}]{kim2023kg}
Jiho Kim, Yeonsu Kwon, Yohan Jo, and Edward Choi. 2023.
\newblock Kg-gpt: A general framework for reasoning on knowledge graphs using large language models.
\newblock In \emph{EMNLP}.

\bibitem[{Kipf and Welling(2016)}]{kipf2016semi}
Thomas~N Kipf and Max Welling. 2016.
\newblock Semi-supervised classification with graph convolutional networks.
\newblock \emph{arXiv}.

\bibitem[{Lazaridou et~al.(2022)Lazaridou, Gribovskaya, Stokowiec, and Grigorev}]{lazaridou2022internet}
Angeliki Lazaridou, Elena Gribovskaya, Wojciech Stokowiec, and Nikolai Grigorev. 2022.
\newblock Internet-augmented language models through few-shot prompting for open-domain question answering.
\newblock \emph{arXiv}.

\bibitem[{Lewis et~al.(2020)Lewis, Perez, Piktus, Petroni, Karpukhin, Goyal, K{\"u}ttler, Lewis, Yih, Rockt{\"a}schel et~al.}]{lewis2020retrieval}
Patrick Lewis, Ethan Perez, Aleksandra Piktus, Fabio Petroni, Vladimir Karpukhin, Naman Goyal, Heinrich K{\"u}ttler, Mike Lewis, Wen-tau Yih, Tim Rockt{\"a}schel, et~al. 2020.
\newblock Retrieval-augmented generation for knowledge-intensive nlp tasks.
\newblock \emph{NeuralIPS}.

\bibitem[{Liu et~al.(2024)Liu, He, Tian, and Chawla}]{liu2024can}
Zheyuan Liu, Xiaoxin He, Yijun Tian, and Nitesh~V Chawla. 2024.
\newblock Can we soft prompt llms for graph learning tasks?
\newblock In \emph{WWW}, pages 481--484.

\bibitem[{Ma et~al.(2025)Ma, Qian, Zhang, Wang, Qian, Bai, Ding, Luo, Zhang, Murugesan et~al.}]{ma2025autodata}
Tianyi Ma, Yiyue Qian, Zheyuan Zhang, Zehong Wang, Xiaoye Qian, Feifan Bai, Yifan Ding, Xuwei Luo, Shinan Zhang, Keerthiram Murugesan, et~al. 2025.
\newblock Autodata: A multi-agent system for open web data collection.
\newblock \emph{arXiv preprint arXiv:2505.15859}.

\bibitem[{MetaAI(2024)}]{meta2024llama3_2_3b_instruct}
MetaAI. 2024.
\newblock Llama-3.2-3b-instruct.
\newblock \url{https://huggingface.co/meta-llama/Llama-3.2-3B-Instruct}.

\bibitem[{Min et~al.(2022)Min, Liu, Xu, and Jiang}]{min2022applications}
Weiqing Min, Chunlin Liu, Leyi Xu, and Shuqiang Jiang. 2022.
\newblock Applications of knowledge graphs for food science and industry.
\newblock \emph{Patterns}.

\bibitem[{MistralAI(2024)}]{mistral2024_mistral7b_instruct_v0_2}
MistralAI. 2024.
\newblock Mistral-7b-instruct-v0.2.
\newblock \url{https://huggingface.co/mistralai/Mistral-7B-Instruct-v0.2}.

\bibitem[{Ni et~al.(2025)Ni, Liu, Wang, Lei, Zhao, Cheng, Zeng, Dong, Xia, Kenthapadi et~al.}]{ni2025towards}
Bo~Ni, Zheyuan Liu, Leyao Wang, Yongjia Lei, Yuying Zhao, Xueqi Cheng, Qingkai Zeng, Luna Dong, Yinglong Xia, Krishnaram Kenthapadi, et~al. 2025.
\newblock Towards trustworthy retrieval augmented generation for large language models: A survey.
\newblock \emph{arXiv}.

\bibitem[{Ong et~al.(2025)Ong, Almahairi, Wu, Chiang, Wu, Gonzalez, Kadous, and Stoica}]{ong2024routellm}
Isaac Ong, Amjad Almahairi, Vincent Wu, Wei‐Lin Chiang, Tianhao Wu, Joseph~E. Gonzalez, M.~Waleed Kadous, and Ion Stoica. 2025.
\newblock Routellm: Learning to route llms with preference data.
\newblock In \emph{ICLR}.

\bibitem[{OpenAI(2024)}]{openai_gpt4o_mini}
OpenAI. 2024.
\newblock Gpt-4o mini: advancing cost-efficient intelligence.
\newblock \url{https://openai.com/index/gpt-4o-mini-advancing-cost-efficient-intelligence/}.

\bibitem[{Peng et~al.(2024)Peng, Zhu, Liu, Bo, Shi, Hong, Zhang, and Tang}]{peng2024graph}
Boci Peng, Yun Zhu, Yongchao Liu, Xiaohe Bo, Haizhou Shi, Chuntao Hong, Yan Zhang, and Siliang Tang. 2024.
\newblock Graph retrieval-augmented generation: A survey.
\newblock \emph{arXiv}.

\bibitem[{Qian et~al.(2025)Qian, Xie, Wang, Liu, Zhu, Xia, Dang, Du, Chen, Yang et~al.}]{qian2025scaling}
Chen Qian, Zihao Xie, YiFei Wang, Wei Liu, Kunlun Zhu, Hanchen Xia, Yufan Dang, Zhuoyun Du, Weize Chen, Cheng Yang, et~al. 2025.
\newblock Scaling large language model-based multi-agent collaboration.
\newblock In \emph{ICLR}.

\bibitem[{Shinn et~al.(2023)Shinn, Cassano, Gopinath, Narasimhan, and Yao}]{shinn2023reflexion}
Noah Shinn, Federico Cassano, Ashwin Gopinath, Karthik Narasimhan, and Shunyu Yao. 2023.
\newblock Reflexion: Language agents with verbal reinforcement learning.
\newblock \emph{NeurIPS}.

\bibitem[{Shirai et~al.(2021)Shirai, Seneviratne, Gordon, Chen, and McGuinness}]{shirai2021identifying}
Sola~S Shirai, Oshani Seneviratne, Minor~E Gordon, Ching-Hua Chen, and Deborah~L McGuinness. 2021.
\newblock Identifying ingredient substitutions using a knowledge graph of food.
\newblock \emph{Frontiers in Artificial Intelligence}.

\bibitem[{Stripelis et~al.(2024)}]{stripelis2024multi}
D.~Stripelis et~al. 2024.
\newblock A multi-model router for efficient llm inference.
\newblock In \emph{EMNLP}.

\bibitem[{Sun et~al.(2019)Sun, Bedrax-Weiss, and Cohen}]{sun2019pullnet}
Haitian Sun, Tania Bedrax-Weiss, and William Cohen. 2019.
\newblock Pullnet: Open domain question answering with iterative retrieval on knowledge bases and text.
\newblock In \emph{EMNLP-IJCNLP}.

\bibitem[{Sun et~al.(2024)Sun, Xu, Tang, Wang, Lin, Gong, Ni, Shum, and Guo}]{sunthink}
Jiashuo Sun, Chengjin Xu, Lumingyuan Tang, Saizhuo Wang, Chen Lin, Yeyun Gong, Lionel Ni, Heung-Yeung Shum, and Jian Guo. 2024.
\newblock Think-on-graph: Deep and responsible reasoning of large language model on knowledge graph.
\newblock In \emph{ICLR}.

\bibitem[{Taunk et~al.(2023)Taunk, Khanna, Kandru, Varma, Sharma, and Tapaswi}]{taunk2023grapeqa}
Dhaval Taunk, Lakshya Khanna, Siri Venkata Pavan~Kumar Kandru, Vasudeva Varma, Charu Sharma, and Makarand Tapaswi. 2023.
\newblock Grapeqa: Graph augmentation and pruning to enhance question-answering.
\newblock In \emph{WWW}.

\bibitem[{Tian et~al.(2022{\natexlab{a}})Tian, Zhang, Guo, Huang, Metoyer, and Chawla}]{tian2022reciperec}
Y~Tian, C~Zhang, Z~Guo, C~Huang, R~Metoyer, and N~Chawla. 2022{\natexlab{a}}.
\newblock Reciperec: A heterogeneous graph learning model for recipe recommendation.
\newblock In \emph{IJCAI}.

\bibitem[{Tian et~al.(2022{\natexlab{b}})Tian, Zhang, Guo, Ma, Metoyer, and Chawla}]{tian2022recipe2vec}
Yijun Tian, Chuxu Zhang, Zhichun Guo, Yihong Ma, Ronald Metoyer, and Nitesh~V Chawla. 2022{\natexlab{b}}.
\newblock Recipe2vec: Multi-modal recipe representation learning with graph neural networks.
\newblock \emph{arXiv}.

\bibitem[{Tian et~al.(2021)Tian, Zhang, Metoyer, and Chawla}]{tian2021recipe}
Yijun Tian, Chuxu Zhang, Ronald Metoyer, and Nitesh~V Chawla. 2021.
\newblock Recipe representation learning with networks.
\newblock In \emph{CIKM}.

\bibitem[{Veli{\v{c}}kovi{\'c} et~al.(2017)Veli{\v{c}}kovi{\'c}, Cucurull, Casanova, Romero, Lio, and Bengio}]{velivckovic2017graph}
Petar Veli{\v{c}}kovi{\'c}, Guillem Cucurull, Arantxa Casanova, Adriana Romero, Pietro Lio, and Yoshua Bengio. 2017.
\newblock Graph attention networks.
\newblock \emph{arXiv}.

\bibitem[{Wang et~al.(2024{\natexlab{a}})Wang, Feng, He, Tan, Han, and Tsvetkov}]{wang2024can}
Heng Wang, Shangbin Feng, Tianxing He, Zhaoxuan Tan, Xiaochuang Han, and Yulia Tsvetkov. 2024{\natexlab{a}}.
\newblock Can language models solve graph problems in natural language?
\newblock \emph{NeuralIPS}.

\bibitem[{Wang et~al.(2025{\natexlab{a}})Wang, Jue, Athiwaratkun, Zhang, and Zou}]{wang2025mixture}
Junlin Wang, WANG Jue, Ben Athiwaratkun, Ce~Zhang, and James Zou. 2025{\natexlab{a}}.
\newblock Mixture-of-agents enhances large language model capabilities.
\newblock In \emph{ICLR}.

\bibitem[{Wang et~al.(2021)Wang, Duan, Jiang, Jing, Song, and Nie}]{wang2021market2dish}
Wenjie Wang, Ling-Yu Duan, Hao Jiang, Peiguang Jing, Xuemeng Song, and Liqiang Nie. 2021.
\newblock Market2dish: health-aware food recommendation.
\newblock \emph{TOMM}.

\bibitem[{Wang et~al.(2025{\natexlab{b}})Wang, Fu, Zhang, Cheng et~al.}]{wang2025mixllm}
X.~Wang, Y.~Fu, Y.~Zhang, W.~Cheng, et~al. 2025{\natexlab{b}}.
\newblock Mixllm: Dynamic routing in mixed large language models.
\newblock In \emph{NAACL}.

\bibitem[{Wang et~al.(2023{\natexlab{a}})Wang, Yang, Qiu, Liang, He, Gu, Xiao, and Wang}]{wang2023knowledgpt}
Xintao Wang, Qianwen Yang, Yongting Qiu, Jiaqing Liang, Qianyu He, Zhouhong Gu, Yanghua Xiao, and Wei Wang. 2023{\natexlab{a}}.
\newblock Knowledgpt: Enhancing large language models with retrieval and storage access on knowledge bases.
\newblock \emph{arXiv}.

\bibitem[{Wang et~al.(2023{\natexlab{b}})Wang, Wei, Schuurmans, Le, Chi, and Zhou}]{wang2023selfconsistency}
Xuezhi Wang, Jason Wei, Dale Schuurmans, Quoc Le, Ed~Chi, and Denny Zhou. 2023{\natexlab{b}}.
\newblock Self-consistency improves chain of thought reasoning in language models.
\newblock In \emph{ICLR}.

\bibitem[{Wang et~al.(2022)Wang, Wei, Schuurmans, Le, Chi, Narang, Chowdhery, and Zhou}]{wang2022self}
Xuezhi Wang, Jason Wei, Dale Schuurmans, Quoc~V Le, Ed~H Chi, Sharan Narang, Aakanksha Chowdhery, and Denny Zhou. 2022.
\newblock Self-consistency improves chain of thought reasoning in language models.
\newblock In \emph{ICLR}.

\bibitem[{Wang et~al.(2025{\natexlab{c}})Wang, Liu, Zhang, Ma, Zhang, and Ye}]{wang2025can}
Zehong Wang, Sidney Liu, Zheyuan Zhang, Tianyi Ma, Chuxu Zhang, and Yanfang Ye. 2025{\natexlab{c}}.
\newblock Can llms convert graphs to text-attributed graphs?
\newblock In \emph{NAACL}.

\bibitem[{Wang et~al.(2024{\natexlab{b}})Wang, Zhang, Chawla, Zhang, and Ye}]{wang2024gft}
Zehong Wang, Zheyuan Zhang, Nitesh Chawla, Chuxu Zhang, and Yanfang Ye. 2024{\natexlab{b}}.
\newblock Gft: Graph foundation model with transferable tree vocabulary.
\newblock \emph{NeruIPS}.

\bibitem[{Wang et~al.(2025{\natexlab{d}})Wang, Zhang, Ma, Chawla, Zhang, and Ye}]{wang2024learning}
Zehong Wang, Zheyuan Zhang, Tianyi Ma, Nitesh~V Chawla, Chuxu Zhang, and Yanfang Ye. 2025{\natexlab{d}}.
\newblock Learning cross-task generalities across graphs via task-trees.
\newblock \emph{ICML}.

\bibitem[{Wang et~al.(2025{\natexlab{e}})Wang, Zhang, Ma, Chawla, Zhang, and Ye}]{wang2025neural}
Zehong Wang, Zheyuan Zhang, Tianyi Ma, Nitesh~V Chawla, Chuxu Zhang, and Yanfang Ye. 2025{\natexlab{e}}.
\newblock Neural graph pattern machine.
\newblock \emph{ICML}.

\bibitem[{Wang et~al.(2024{\natexlab{c}})Wang, Zhang, Zhang, and Ye}]{wang2024subgraph}
Zehong Wang, Zheyuan Zhang, Chuxu Zhang, and Yanfang Ye. 2024{\natexlab{c}}.
\newblock Subgraph pooling: tackling negative transfer on graphs.
\newblock In \emph{IJCAI}.

\bibitem[{Wei et~al.(2022)Wei, Wang, Schuurmans, Bosma, Xia, Chi, Le, Zhou et~al.}]{wei2022chain}
Jason Wei, Xuezhi Wang, Dale Schuurmans, Maarten Bosma, Fei Xia, Ed~Chi, Quoc~V Le, Denny Zhou, et~al. 2022.
\newblock Chain-of-thought prompting elicits reasoning in large language models.
\newblock \emph{NeuralIPS}.

\bibitem[{Wen et~al.(2023)Wen, Wang, and Sun}]{wen2023mindmap}
Yilin Wen, Zifeng Wang, and Jimeng Sun. 2023.
\newblock Mindmap: Knowledge graph prompting sparks graph of thoughts in large language models.
\newblock \emph{arXiv}.

\bibitem[{WHO(2021)}]{WHO_2021_HealthyDiet}
WHO. 2021.
\newblock \href {https://www.who.int/news-room/fact-sheets/detail/healthy-diet} {Healthy diet}.

\bibitem[{WHO(2023)}]{WHO_2023_Obesity}
WHO. 2023.
\newblock \href {https://www.who.int/health-topics/obesity} {Obesity info page of the world health organization}.

\bibitem[{Yao et~al.(2023)Yao, Zhao, Yu, Du, Shafran, Narasimhan, and Cao}]{yao2023react}
Shunyu Yao, Jeffrey Zhao, Dian Yu, Nan Du, Izhak Shafran, Karthik Narasimhan, and Yuan Cao. 2023.
\newblock React: Synergizing reasoning and acting in language models.
\newblock In \emph{ICLR}.

\bibitem[{Yasunaga et~al.(2021)Yasunaga, Ren, Bosselut, Liang, and Leskovec}]{yasunaga2021qa}
Michihiro Yasunaga, Hongyu Ren, Antoine Bosselut, Percy Liang, and Jure Leskovec. 2021.
\newblock Qa-gnn: Reasoning with language models and knowledge graphs for question answering.
\newblock In \emph{NAACL}.

\bibitem[{Yue et~al.(2021)Yue, Wang, Zhang, and Liu}]{yue2021overview}
Wenbin Yue, Zidong Wang, Jieyu Zhang, and Xiaohui Liu. 2021.
\newblock An overview of recommendation techniques and their applications in healthcare.
\newblock \emph{IEEE/CAA Journal of Automatica Sinica}.

\bibitem[{Zhang et~al.(2022)Zhang, Zhang, Yu, Tang, Tang, Li, and Chen}]{zhang2022subgraph}
Jing Zhang, Xiaokang Zhang, Jifan Yu, Jian Tang, Jie Tang, Cuiping Li, and Hong Chen. 2022.
\newblock Subgraph retrieval enhanced model for multi-hop knowledge base question answering.
\newblock In \emph{ACL}.

\bibitem[{Zhang et~al.(2024{\natexlab{a}})Zhang, Zhang, Zhou, and Shen}]{zhang2024greenrec}
Lingzi Zhang, Yinan Zhang, Xin Zhou, and Zhiqi Shen. 2024{\natexlab{a}}.
\newblock Greenrec: A large-scale dataset for green food recommendation.
\newblock In \emph{WWW}.

\bibitem[{Zhang et~al.(2025{\natexlab{a}})Zhang, Ge, Li, Zhu, Zhang, and Ye}]{zhang2025mapro}
Zheyuan Zhang, Lin Ge, Hongjiang Li, Weicheng Zhu, Chuxu Zhang, and Yanfang Ye. 2025{\natexlab{a}}.
\newblock \href {https://arxiv.org/abs/2510.07475} {Mapro: Recasting multi-agent prompt optimization as maximum a posteriori inference}.
\newblock \emph{arXiv}.

\bibitem[{Zhang et~al.(2025{\natexlab{b}})Zhang, Li, Le, Wang, Ma, Galassi, Murugesan, Moniz, Geyer, Chawla et~al.}]{zhang2024ngqa}
Zheyuan Zhang, Yiyang Li, Nhi Ha~Lan Le, Zehong Wang, Tianyi Ma, Vincent Galassi, Keerthiram Murugesan, Nuno Moniz, Werner Geyer, Nitesh~V Chawla, et~al. 2025{\natexlab{b}}.
\newblock Ngqa: A nutritional graph question answering benchmark for personalized health-aware nutritional reasoning.
\newblock \emph{ACL}.

\bibitem[{Zhang et~al.(2025{\natexlab{c}})Zhang, Ma, Wang, Li, Hou, Sun, Shi, Ma, Song, Abbasi et~al.}]{zhang2025llms4all}
Zheyuan Zhang, Tianyi Ma, Zehong Wang, Yiyang Li, Shifu Hou, Weixiang Sun, Kaiwen Shi, Yijun Ma, Wei Song, Ahmed Abbasi, et~al. 2025{\natexlab{c}}.
\newblock Llms4all: A review on large language models for research and applications in academic disciplines.
\newblock \emph{arXiv preprint arXiv:2509.19580}.

\bibitem[{Zhang et~al.(2025{\natexlab{d}})Zhang, Shi, Yuan, Wang, Ma, Murugesan, Galassi, Zhang, and Ye}]{zhang2025agentrouter}
Zheyuan Zhang, Kaiwen Shi, Zhengqing Yuan, Zehong Wang, Tianyi Ma, Keerthiram Murugesan, Vincent Galassi, Chuxu Zhang, and Yanfang Ye. 2025{\natexlab{d}}.
\newblock Agentrouter: A knowledge-graph-guided llm router for collaborative multi-agent question answering.
\newblock \emph{arXiv}.

\bibitem[{Zhang et~al.(2024{\natexlab{b}})Zhang, Wang, Hou, Hall, Bachman, White, Galassi, Chawla, Zhang, and Ye}]{zhang2024diet}
Zheyuan Zhang, Zehong Wang, Shifu Hou, Evan Hall, Landon Bachman, Jasmine White, Vincent Galassi, Nitesh~V Chawla, Chuxu Zhang, and Yanfang Ye. 2024{\natexlab{b}}.
\newblock Diet-odin: A novel framework for opioid misuse detection with interpretable dietary patterns.
\newblock In \emph{Proceedings of the 30th ACM SIGKDD Conference on Knowledge Discovery and Data Mining}, pages 6312--6323.

\bibitem[{Zhang et~al.(2024{\natexlab{c}})Zhang, Wang, Ma, Taneja, Nelson, Le, Murugesan, Ju, Chawla, Zhang et~al.}]{zhang2024mopi}
Zheyuan Zhang, Zehong Wang, Tianyi Ma, Varun~Sameer Taneja, Sofia Nelson, Nhi Ha~Lan Le, Keerthiram Murugesan, Mingxuan Ju, Nitesh~V Chawla, Chuxu Zhang, et~al. 2024{\natexlab{c}}.
\newblock Mopi-hfrs: A multi-objective personalized health-aware food recommendation system with llm-enhanced interpretation.
\newblock \emph{arXiv}.

\bibitem[{Zhong et~al.(2024)Zhong, Wang, and Shang}]{zhong2024debug}
Li~Zhong, Zilong Wang, and Jingbo Shang. 2024.
\newblock Debug like a human: A large language model debugger via verifying runtime execution step by step.
\newblock In \emph{Findings of ACL}.

\end{thebibliography}

\newpage
\appendix

\section{Additional Related Works}
\label{app:literature}
\subsection{Knowledge Graph Question Answering}
Research on Knowledge Graph Question Answering (KGQA) has progressed from classic semantic parsing and retrieval paradigms to increasingly model-driven solutions. Early systems translated natural-language questions into executable logical forms (e.g., SPARQL) over a knowledge graph \cite{sun2019pullnet, zhang2022subgraph}, often pairing pre-trained encoders such as BERT with graph-aware architectures (GNNs/LSTMs) to locate entities, relations, and supporting subgraphs \cite{yasunaga2021qa, taunk2023grapeqa}.  
More recent approaches incorporate large language models (LLMs) to improve both access and reasoning: some convert questions into structured queries like SQL/SPARQL to sharpen retrieval \cite{jiang2023structgpt, wang2023knowledgpt}, while others emphasize multi-hop inference over retrieved triples or subgraphs to handle compositional reasoning \cite{kim2023kg, gao2024two}. Despite these advances, widely used benchmarks remain largely general-purpose and do not fully capture domain-specific demands—e.g., the nuanced constraints present in nutritional-health reasoning scenarios.

\subsection{Graph-Retrieval Augmented Generation}
Graph-Retrieval Augmented Generation (Graph-RAG) generalizes the RAG paradigm \cite{lewis2020retrieval, ni2025towards} by retrieving \emph{structured} evidence rather than only unstructured text. Instead of passages alone, Graph-RAG surfaces graph fragments (triples/subgraphs) and uses graph encoders to condition generation, thereby improving precision and reducing redundancy \cite{guo2024knowledgenavigator, wen2023mindmap, lazaridou2022internet, liu2024can}.  
Current evaluations predominantly probe elementary graph reasoning skills—such as path finding, degree/counting, or edge existence \cite{fatemi2023talk, wang2024can, wang2025can}. While informative for fundamentals, these settings under-represent domain-specific requirements. He et al.\ introduce more advanced graph-understanding benchmarks in general contexts \cite{he2024g}. Building on Graph-RAG principles, many applications nowadays thrive and shed lights on new research paths \cite{zhang2024diet}.

\subsection{Graph Neural Networks.}
Graph Neural Networks (GNNs) are designed for relational data and have delivered strong results across social, recommendation, biological, and molecular applications by exploiting graph inductive biases \cite{kipf2016semi, velivckovic2017graph, hamilton2017inductive}. Their ability to share parameters across varying graph sizes/topologies supports deployment in dynamic, real-world settings. A growing body of work investigates transfer and pretraining for cross-task/domain generalization—mirroring trends in language and vision—via subgraph pooling, pretraining schemes, and task-agnostic embeddings \cite{wang2024subgraph, cao2023pre}. Looking forward, the community is moving toward \emph{graph foundation models}, i.e., large-scale pretrained GNN backbones intended to capture broadly reusable structural/semantic patterns \cite{wang2024gft, wang2025neural, wang2024learning}. Despite progress, open challenges persist, including oversmoothing, expressive-power limits, and scalability, motivating research on more adaptive architectures and training recipes.

\begin{table}[!t]
    \centering
    \setlength{\tabcolsep}{10pt} % Adjust column separation
    \resizebox{\linewidth}{!}{
        \begin{tabular}{lccc}
            \toprule
            \textbf{Nutrients} & \textbf{Low Threshold} & \textbf{High Threshold} & \textbf{NRV} \\
            \midrule\midrule
            Calories (kcal)   & 40   & 225   & 2000 \\
            Carbohydrates (g)  & 55   & 75    & -    \\
            Protein (g)     & 10   & 15    & 50   \\
            Saturated Fat (g)   & 1.5  & 5     & 20   \\
            Cholesterol (mg)    & 20   & 40    & 300  \\
            Sugar (g)           & 5    & 22.5  & -    \\
            Dietary Fiber (g)    & 3    & 6     & -    \\
            \midrule\midrule
            Sodium (mg)        & 120  & 200   & 2000 \\
            Potassium (mg)       & 0    & 525   & 3500 \\
            Phosphorus (mg)      & 0    & 105   & 700  \\
            Iron (mg)            & 0    & 3.3   & 22   \\
            Calcium (mg)         & 0    & 150   & 1000 \\
            Folic Acid (µg)      & 0    & 60    & 400  \\
            Vitamin C (mg)       & 0    & 15    & 100  \\
            Vitamin D (µg)       & 0    & 2.25  & 15   \\
            Vitamin B12 (µg)     & 0    & 0.36  & 2.4  \\
            \bottomrule
        \end{tabular}
    }
    \caption{Nutrient Reference Values (NRV) and thresholds (per 100g of food) used based on the nutritional standards.}
    \vspace{-10pt}
    \label{tab:nutrient_thresholds}
\end{table}

\begin{table}[!t]
    \centering
    \setlength{\tabcolsep}{8pt} % Adjust column separation
    \resizebox{\linewidth}{!}{
        \begin{tabular}{lcc}
            \toprule
            \textbf{Health Indicator} & \textbf{High Threshold} & \textbf{Low Threshold} \\
            \midrule\midrule
            BMI                         & 30           & 18.5         \\
            Waist Circumference (cm)    & 102 (88)     & -            \\
            Blood Pressure (mmHg)       & 140          & 90           \\
            Osteoporosis                & -            & -            \\
            \midrule
            Blood Urea Nitrogen (mmol/L) & 7.1         & -            \\
            Low-Density Lipoprotein (mmol/L) & 3.3      & -            \\
            Red Blood Cell (million cells/uL) & -       & 4            \\
            Glucose (mmol/L)            & 7            & -            \\
            Glycohemoglobin (\%)        & 6.5          & -            \\
            Hemoglobin (g/dL)           & -            & 13.2 (11.6)  \\
            \bottomrule
        \end{tabular}
    }
    \caption{Health Indicators with Corresponding High and Low Thresholds. Parentheses indicate sex-specific: male (female) thresholds where applicable.}
    \vspace{-10pt}
    \label{tab:health_indicators}
\end{table}

\section{NGQA Benchmark Details}
\label{app:benchmarks}

\subsection{Benchmark Description}
We build upon the NGQA benchmark \cite{zhang2024ngqa}, which integrates multiple U.S. national health and nutrition resources. The benchmark primarily relies on the National Health and Nutrition Examination Survey (NHANES), a biannual survey conducted by the CDC that combines demographic, dietary, examination, laboratory, and questionnaire data. Dietary intake is captured through the What We Eat in America (WWEIA) program, which collects 24-hour dietary recalls and links them to nutrient information in the USDA’s Food and Nutrient Database for Dietary Studies (FNDDS). Together, these resources provide comprehensive coverage of health indicators, food consumption, and nutrient composition. To represent user behaviors, the benchmark includes dietary habit features derived from NHANES (e.g., awareness of healthy eating, consumption of processed or frozen foods). Through manual curation, they constructed 54 distinct dietary habit tags, which serve as additional nodes in the graph.

Nutrient annotations were derived from international dietary standards (WHO, FSA, EU, CAC), covering 16 macro- and micronutrients such as calories, protein, sugar, sodium, and iron. These were mapped to user health profiles through threshold-based rules with mapping rules shown in Table~\ref{tab:nutrient_thresholds}, \ref{tab:health_indicators} and \ref{tab:health_indicators_tags}. The benchmark emphasizes four prevalent health conditions with clear dietary relevance: obesity, hypertension, diabetes, and opioid misuse. Standard clinical definitions were applied to the first three, while opioid misuse was defined using medical criteria based on illicit use or long-term prescription opioid records.

\begin{table}[!t]
    \centering
    \setlength{\tabcolsep}{8pt} % Adjust column separation
    \resizebox{\linewidth}{!}{ % Fit to single-column width
        \begin{tabular}{ll}
            \toprule
            \textbf{Health Indicator} & \textbf{Associated Tags} \\
            \midrule
            Obesity                         & Low Calorie \\
            Opioid Misuse                   & High Protein; Low Sugar; Low Sodium \\
            Hypertension                    & Low Sodium \\
            Diabetes                        & Low Sugar; Low Carb \\
            Weight Loss/Low Calorie Diet    & Low Calorie \\
            Low Fat/Low Cholesterol Diet    & Low Cholesterol; Low Saturated Fat \\
            Low Salt/Low Sodium Diet        & Low Sodium \\
            Sugar-Free/Low Sugar Diet       & Low Sugar \\
            Diabetic Diet                   & Low Sugar; Low Carb \\
            Weight Gain/Muscle Building Diet & High Calorie; High Protein \\
            Low Carbohydrate Diet           & Low Carb \\
            High Protein Diet               & High Protein \\
            Renal/Kidney Diet               & Low Protein \\
            \bottomrule
        \end{tabular}
    }
    \caption{Health Indicators and Their Associated Nutritional Tags. Each indicator is linked to relevant tags reflecting dietary requirements.}
    \vspace{-10pt}
    \label{tab:health_indicators_tags}
\end{table}

\subsection{Datasets}  
Building on the dataset foundation, the NGQA benchmark organizes its tasks into three different datasets, each reflecting different levels of information availability and reasoning complexity.  

\noindent\textbf{Sparse Dataset.} These address scenarios with minimal information. Each food is linked to only one nutrition tag associated with a single user health condition. While this setting mirrors real-world cases where labels are scarce or incomplete, it presents substantial challenges: the one-to-one linkage amplifies the difficulty of subgraph retrieval and makes models especially susceptible to interference from irrelevant nodes.  

\noindent\textbf{Standard Dataset.} These represent the balanced and idealized setting of the benchmark. Foods are connected to multiple nutrition tags that either match or contradict several user health conditions. Such cases capture clear-cut relationships between dietary choices and health outcomes, providing a controlled environment for evaluating model performance. Standard questions therefore serve as the reference baseline for structured reasoning tasks.  

\noindent\textbf{Complex Dataset.} These replicate the intricacies of real-life nutritional decision-making. Foods may simultaneously contain tags that both align with and conflict with user health conditions. For example, a food might be low in sodium (beneficial for hypertension) but high in sugar (problematic for diabetes). Models must navigate these conflicting signals, prioritize health needs, and perform trade-off reasoning, making this category the most realistic and challenging.  

\noindent A statistical breakdown of these three categories is provided in Table~\ref{tab:benchmark_stats}. To further quantify informativeness, the benchmark also introduces a Signal-to-Noise Ratio (SNR), defined as the ratio of relevant nodes or tags (\textit{signal}) to the total number of nodes or tags in the graph (\textit{noise}). As shown in Table~\ref{tab:snr_stats}, sparse questions have the lowest SNR, reflecting their limited resources, while complex questions achieve the highest SNR, highlighting the richer contextual information needed for better reasoning.

\begin{table}[!t]
    \centering
    \setlength{\tabcolsep}{6pt} % Adjust column separation for tighter fit
    \resizebox{\columnwidth}{!}{ % Resize to fit the column width
        \begin{tabular}{lccc}
            \toprule
            \textbf{Question Level} & \textbf{\# Records} & \textbf{Avg. \# Nodes} & \textbf{Avg. \# Edges} \\
            \midrule\midrule
            Sparse      & 8,490 & 25.8 & 24.9  \\
            Standard    & 3,622 & 28.2 & 29.0  \\
            Complex     & 1,690 & 30.9 & 34.0  \\
            \bottomrule
        \end{tabular}
    }
    \vspace{-5pt}
    \caption{Statistics of the Benchmark by Question Level.}
    \vspace{-10pt}
    \label{tab:benchmark_stats}
\end{table}

\begin{table}[!t]
    \centering
    \setlength{\tabcolsep}{6pt} % Adjust column separation for tighter fit
    \resizebox{\columnwidth}{!}{ % Resize to fit the column width
        \begin{tabular}{lcc}
            \toprule
            \textbf{Question Level} & \textbf{Avg. Node SNR} & \textbf{Avg. Tag SNR} \\
            \midrule\midrule
            Sparse      & 16.4 & 19.3 \\
            Standard    & 24.7 & 49.4 \\
            Complex     & 31.6 & 76.3 \\
            \bottomrule
        \end{tabular}
    }
    \vspace{-5pt}
    \caption{Signal-to-Noise Ratio (SNR) by Question Level.}
    \vspace{-10pt}
    \label{tab:snr_stats}
\end{table}

\section{Baseline and Implementation Details}
\label{app:baselines}

To cover the range of agentic strategies explored in contemporary LLM research, the benchmark incorporates six representative designs. \textit{Raw} serves as the direct prompting baseline, reflecting the unaugmented capacity of the backbone model. \textit{Chain-of-Thought (CoT)}~\citep{wei2022chain} encourages models to generate intermediate reasoning steps, thereby improving performance on multi-step tasks. \textit{Self-Consistency (SC)}~\citep{wang2023selfconsistency} extends CoT by sampling multiple reasoning paths and aggregating the most consistent outcome, reducing sensitivity to individual trajectories. \textit{ReAct-Reflection}~\citep{yao2023react, shinn2023reflexion} integrates reasoning with external actions and iterative self-correction, yielding more grounded and robust responses. \textit{Multi-Agent Debate (MAD)}~\citep{du2024mad} facilitates adversarial discussion among agents, enabling consensus through structured deliberation. Finally, \textit{Multi-Agent Summary} aggregates partial reasoning produced by diverse agents into a distilled final answer. Together, these designs span from single-agent prompting to multi-agent collaboration, underscoring the need for principled routing across heterogeneous strategies~\citep{zhang2025mapro}.  

Beyond heuristics, we benchmark against three state-of-the-art LLM routing frameworks. \textbf{LLM-Blender}~\citep{jiang2023llm} leverages an LLM-as-a-judge paradigm: candidate outputs from multiple agents are presented to a meta-LLM, which adjudicates and selects the final answer. This approach highlights the potential of reflective meta-reasoning but incurs high cost and latency due to repeated LLM calls. \textbf{HybridLLM}~\citep{ding2024hybridllm} adopts a hybrid strategy that combines lightweight scoring heuristics with selective meta-LLM adjudication, aiming to balance efficiency and effectiveness. We also select two baselines reported in NGQA paper, which can be seen as classical graph retrieval augmented methods. \textbf{KAPING}~\citep{baek2023knowledge} answers questions by constructing a subgraph from the entities mentioned in the query and their neighbors, which is then linearized into triples and fed into the LLM. While the original implementation applies top-$k$ filtering, this step was omitted in our benchmark as it is not applicable when only user and food entities are present. \textbf{ToG}~\citep{sunthink} iteratively explores and prunes reasoning paths on a knowledge graph to identify candidate answers. Since the released code is tailored to Wikidata/Freebase, we reimplemented ToG for our benchmark and introduced two adjustments: increasing the pruning width to five paths and delaying pruning until the second iteration. These modifications ensure sufficient coverage and better alignment with the benchmark’s complexity.  

Beside the implemented baselines, it is worth noting our work is greatly inspired by the broader idea of graph-based reasoning for LLM orchestration in GraphRouter \cite{feng2024graphrouter}. However, our work departs fundamentally from GraphRouter in many ways, of which the most important one is that GraphRouter encodes a query and its context as a single node linked to candidate LLMs, which doesn't fit into our problem settings.

\section{Case Studies}
We provide two case studies to demonstrate the effectiveness of our model in routing questions to the most appropriate agents (Figure~\ref{fig:agent-routing-three-examples}). In the \textit{Borscht} example, the gold answer is \emph{low\_sugar, low\_sodium}. The agent probability distribution shows that \texttt{Qwen2.5-7B-Instruct::raw} and \texttt{Qwen2.5-7B-Instruct::cot} achieve probabilities above 0.08 and both predict the correct labels. Other agents, including those from different backbones such as \texttt{Mistral-7B-Instruct}, also provide correct or near-correct answers, which demonstrates that our router not only identifies the most relevant agents but also assigns higher probabilities to those generating correct predictions. This reduces the chance of routing errors and improves overall reliability in multi-agent collaboration.

We illustrate several examples of applying our subgraph retrieval approach to different datasets in Figure~\ref{fig:entity-importance-three-examples}. To make the reasoning process easier to understand, we first convert the original graph into plain text. The graph for the \textit{Bruschetta} example shows that the food is associated with various ingredients (garlic, olive oil, tomatoes, basil, salt, etc.), belongs to categories like "vegetable sandwiches/burgers," and is linked to multiple nutrition tags such as \textit{low\_carb}, \textit{low\_sugar}, and \textit{high\_sodium}. It also contains user-related information such as health conditions (\textit{hypertension}) and lifestyle habits (e.g., "drinks alcohol less than average," "takes more supplements"). While this graph contains 25 nodes and 30 edges, many of these nodes are not directly useful for answering the question about nutrient tags. Most of the user habits and ingredient details serve as noise that could mislead the reasoning process. Our scoring method solves this problem by assigning importance scores to all nodes. Irrelevant information typically receives scores below 0.01, while truly relevant entities such as \textit{Bruschetta}, \textit{hypertension}, and \textit{high\_sodium} achieve much higher scores. By filtering the graph to include only nodes with scores above 0.01, we obtain a clearer and more focused representation, reducing noise and helping the agent concentrate on the key information needed to answer the question.

\begin{figure*}[t]
\centering
\begin{tcolorbox}[colback=myblue!5!white,
                  colframe=myblue!75!black,
                  title=Agent Routing Case Study]
{

\textbf{Example 1}\\
\textbf{Question:} Based on the nutrients the food provides and the user needs, please answer what nutrient tags are used to determine whether the food "Broccoli cheese soup, prepared with milk, home recipe, canned, or ready-to-serve" is healthy or unhealthy for the user?\\
\textbf{Gold Answer:} low\_protein

\vspace{4pt}
\textbf{Agent probs}\\
\texttt{GPT-4o-mini::AGENT::raw} $\rightarrow$ \emph{low\_protein} (0.052638724)\\
\texttt{GPT-4o-mini::AGENT::cot} $\rightarrow$ \emph{high\_sodium, low\_protein} (0.050060976)\\
\texttt{GPT-4o-mini::AGENT::react\_reflect}$\rightarrow$ \emph{high\_sodium, low\_protein} (0.049231380)\\
\texttt{Mistral-7B-Instruct::AGENT::react\_reflect} $\rightarrow$ \emph{high\_sodium, low\_protein} (0.044603113)\\
\dots\\
\texttt{Mistral-7B-Instruct::AGENT::mad} $\rightarrow$ \emph{low\_carb, low\_sugar, high\_sodium} (0.042540662)\\
\texttt{Llama-3.2-3B::AGENT::cot} $\rightarrow$ \emph{low\_saturated\_fat, low\_cholesterol, low\_carb} (0.039508183)\\
\texttt{Qwen2.5-7B-Instruct::AGENT::cot} $\rightarrow$ \emph{low\_carb, low\_sugar, high\_sodium} (0.032161828)\\
\texttt{Qwen2.5-7B-Instruct::AGENT::raw} $\rightarrow$ \emph{low\_carb, low\_sugar, high\_sodium} (0.031613167)\\

\vspace{10pt}
\textbf{Example 2}\\
\textbf{Question:} Based on the nutrients the food provides and the user needs, please answer what nutrient tags are used to determine whether the food "Borscht" is healthy or unhealthy for the user?\\
\textbf{Gold Answer:} low\_sugar, low\_sodium

\vspace{4pt}
\textbf{Agent probs}\\
\texttt{Qwen2.5-7B-Instruct::AGENT::raw} $\rightarrow$ \emph{low\_sugar, low\_sodium} (0.095538996)\\
\texttt{Qwen2.5-7B-Instruct::AGENT::cot} $\rightarrow$ \emph{low\_sugar, low\_sodium} (0.086759798)\\
\texttt{Qwen2.5-7B-Instruct::AGENT::summary} $\rightarrow$ \emph{low\_sugar, low\_sodium} (0.061009243)\\
\texttt{Mistral-7B-Instruct::AGENT::mad} $\rightarrow$ \emph{low\_sugar, low\_sodium} (0.051758543)\\
\dots\\
\texttt{Llama-3.2-3B::AGENT::cot} $\rightarrow$ \emph{low\_sugar, low\_sodium} (0.020628655)\\
\texttt{GPT-4o-mini::AGENT::mad} $\rightarrow$ \emph{low\_carb, low\_sugar, low\_calorie, low\_protein, low\_cholesterol, low\_saturated\_fat, low\_sodium} (0.019520836)\\
\texttt{GPT-4o-mini::AGENT::sc} $\rightarrow$ \emph{low\_carb, low\_sugar, low\_calorie, low\_protein, low\_cholesterol, low\_saturated\_fat, low\_sodium} (0.018208018)\\
\texttt{GPT-4o-mini::AGENT::cot} $\rightarrow$
\emph{low\_carb, low\_sugar, low\_calorie, low\_protein, low\_cholesterol, low\_saturated\_fat, low\_sodium} (0.011033830)\\

}
\end{tcolorbox}

\vspace{-8pt}
\caption{Per-question agent routing cases. Each case shows the question, gold answer, and agents' probability with their generated answers.}
\label{fig:agent-routing-three-examples}
\end{figure*}

\begin{figure*}[t]
\centering
\begin{tcolorbox}[colback=myblue!5!white,
                  colframe=myblue!75!black,
                  title=Subgraph Retrieval Case Study]
{

\textbf{Case 1 (Standard)}\\
\textbf{Graph Context}:\\
\footnotesize
\texttt{'Biscayne codfish, Puerto Rican style', 'belongs to', 'low\_carb'], ['Biscayne codfish, Puerto Rican style', 'belongs to', 'low\_sugar'], ['Biscayne codfish, Puerto Rican style', 'belongs to', 'high\_sodium'], ['user', 'has', 'diabetes'], ['diabetes', 'match', 'low\_sugar'], ['obesity', 'need', 'low\_calorie'], \ldots}
\normalsize

\textbf{Node Quantity}: 31\\
\textbf{edge quantity:} 31

\textbf{Question:} Based on the nutrients the food provides and the user needs, please answer what nutrient tags are used to determine whether the food "Biscayne codfish, Puerto Rican style" is healthy or unhealthy for the user?\\
\textbf{Gold Answer:} low\_carb, low\_sugar

\textbf{Entity Importance}\\
user (0.412504), low\_carb (0.210431), Biscayne codfish, Puerto Rican style (0.176805), 
low\_sugar (0.142012),  diabetes (0.025764),  high\_sodium (0.002139), low\_cholesterol (0.002011), \ldots

\vspace{10pt}
\textbf{Case 2 (Complex)}\\
\textbf{Graph Context}:\\
\footnotesize
\texttt{['Matzo ball soup', 'belongs to', 'low\_carb'], ['Matzo ball soup', 'belongs to', 'low\_sugar'], ['Matzo ball soup', 'belongs to', 'high\_sodium'], ['user', 'has', 'diabetes'], ['diabetes', 'match', 'low\_sugar'], ['High protein diet', 'contradict', 'low\_protein'],  \ldots}
\normalsize

\textbf{Node Quantity:} 28\\
\textbf{edge quantity:} 30

\textbf{Question:} Based on the nutrients the food provides and the user needs, please answer what nutrient tags are used to determine whether the food "Matzo ball soup" is healthy or unhealthy for the user?\\
\textbf{Gold Answer:} low\_carb, low\_sugar, low\_protein

\textbf{Entity Importance}\\
Matzo ball soup (0.397344), user (0.168014), low\_protein (0.104383), low\_sugar (0.102992), low\_carb (0.099755), diabetes (0.016299), 
Low carbohydrate diet (0.013385), High protein diet (0.012389), Adds little to no salt at table (0.005626), Eats lots of fish (0.005585), \ldots

\vspace{10pt}
\textbf{Case 3 (Sparse)}\\
\textbf{Graph Context}:\\
\footnotesize
\texttt{['Bruschetta', 'belongs to', 'high\_sodium'], ['Bruschetta', 'has', 'Salt, table, iodized'], ['Bruschetta', 'has', 'Olive oil'], ['user', 'has', 'hypertension'], ['hypertension', 'contradict', 'high\_sodium'], \ldots}
\normalsize

\textbf{Node Quantity:} 25\\
\textbf{edge quantity:} 30

\medskip
\textbf{Question:} Based on the nutrients the food provides and the user needs, please answer what nutrient tags are used to determine whether the food "Bruschetta" is healthy or unhealthy for the user?\\
\textbf{Gold Answer:} high\_sodium

\vspace{4pt}
\textbf{Entity Importance}\\
user (0.283775), Bruschetta (0.283627), hypertension (0.188398), 
high\_sodium (0.187762), \\
Drinks Alcohol less than average (0.003046), Takes more supplements (0.003046), \ldots

}
\end{tcolorbox}

\vspace{-8pt}
\caption{Per-question cases for subgraph retrieval. Each case shows the graph context, the question, the gold answer, and entities with importance score $>$\,0.01; for each listed entity we also include the next two entities to highlight the score cliff (remaining tail elided with "\ldots").}
\label{fig:entity-importance-three-examples}
\end{figure*}

\section{Prompt Designs}
To demonstrate the exact instructions used in our system, we present the full set of prompts that guided the different agent roles. Figures~\ref{fig:qa-prompts} provide a complete overview. These prompts are not intended as a novel design contribution, but rather as transparent documentation of the configurations employed in our experiments.

\begin{figure*}[t]
\centering
\begin{tcolorbox}[colback=myred!5!white,                
                  colframe=myred!75!black,
                  title=QA/Reasoning Prompt Suite]
{

\textbf{raw:}\\
Given a question, a news context, and retrieved documents, answer the question directly.\\
Compress your answer into the SHORTEST exact entity only.\\
The final output must be one JSON object: \texttt{{"answer": "<short factual answer>"}}.

\vspace{6pt}
\textbf{cot (chain-of-thought):}\\
You are a multi-hop reasoning expert and QA agent.\\
Given a question and the context, reason step-by-step before answering.\\
Compress your answer into the SHORTEST exact entity only.\\
The final output must be one JSON object: \texttt{{"answer": "<short factual answer>"}}.

\vspace{6pt}
\textbf{sc (self-consistency):}\\
You are a self-consistency agent. Independently sample multiple plausible entity selections for the given question and context,\\
then internally perform majority voting to decide the final set.\\
Generate diverse candidate sets internally, then pick the majority-agreed entities.\\
The final output must be one JSON object: \texttt{{"answer": "<short factual answer>"}}.

\vspace{6pt}
\textbf{mad (multi-agent debate):}\\
You simulate three roles: \emph{debate\_debater\_a}, \emph{debate\_debater\_b}, and \emph{debate\_judge}.\\
\hspace*{1em}- Debater A proposes the most plausible answer using only the provided context, supported by 1--3 short quotes.\\
\hspace*{1em}- Debater B stress-tests A’s claim: if weak or incomplete, correct it or propose a better alternative.\\
\hspace*{1em}- Debate Judge decides the best final answer using only the given context (noun, number, or yes/no).\\
Finally, Debate Judge condenses the result into the shortest exact entity only.\\
The final output must be one JSON object: \texttt{{"answer": "<short factual answer>"}}.

\vspace{6pt}
\textbf{react\_reflect:}\\
You simulate two roles: \emph{react} and \emph{reflect}.\\
\hspace*{1em}- React is a multi-hop reasoning expert that chains facts into a reasoning plan and derives a brief final answer.\\
\hspace*{1em}- Reflect evaluates React’s answer. If incorrect or incomplete, provide revision suggestions; otherwise, confirm correctness.\\
Reflect then condenses the result into the shortest exact entity only.\\
The final output must be one JSON object: \texttt{{"answer": "<short factual answer>"}}.

\vspace{6pt}
\textbf{summary:}\\
You simulate three roles: \emph{think\_a}, \emph{think\_b}, and \emph{summarize}.\\
\hspace*{1em}- Think\_a and Think\_b independently reason step-by-step and produce candidate answers.\\
\hspace*{1em}- Summarize compares their outputs: if they agree, return the shared answer; if not, select the best one with reasoning.\\
Finally, Summarize condenses the result into the shortest exact entity only.\\
The final output must be one JSON object: \texttt{{"answer": "<short factual answer>"}}.

}
\end{tcolorbox}

\vspace{-10pt}
\caption{Prompt suite for six major agent roles used in NGRouter. Each prompt defines a distinct reasoning behavior that collectively improves multi-agent QA performance.}
\label{fig:qa-prompts}
\end{figure*}

\end{document}